\documentclass[twocolumn]{svjour3}         
\smartqed
\usepackage{graphicx}
\usepackage{amsmath}
\usepackage{array}
\usepackage{booktabs}
\usepackage{algpseudocode}
\usepackage{caption}
\usepackage{tabularx}
\usepackage{threeparttable}
\usepackage{dcolumn}
\usepackage{pifont}
\usepackage[colorlinks,linkcolor=red]{hyperref}
\usepackage{algorithm}
\usepackage{algorithmicx}
\usepackage{subfigure}
\usepackage{multicol}
\usepackage{multirow}
\usepackage{amssymb}
\usepackage{bm}
\usepackage{mathptmx}

\newcolumntype{.}{D{.}{.}{-1}}

\makeatletter
\def\hlinew#1{%
  \noalign{\ifnum0=`}\fi\hrule \@height #1 \futurelet
   \reserved@a\@xhline}
\makeatother

\begin{document}

\title{EdgeStereo: An Effective Multi-Task Learning Network for Stereo Matching and Edge Detection
}

\titlerunning{EdgeStereo: An Effective Multi-Task Learning Network for Stereo Matching and Edge Detection}

\author{Xiao Song \and Xu Zhao* \and Liangji Fang \and Hanwen Hu}

\authorrunning{Xiao Song, Xu Zhao, Liangji Fang and Hanwen Hu}

\institute{* Corresponding author. Authors are with the Department of Automation, Shanghai Jiao Tong University, China. E-mail: \{song\_xiao,zhaoxu,fangliangji,huhanwen\}@sjtu.edu.cn.}

\date{Received: 8 March 2019 / Accepted: 10 December 2019}

\maketitle

\begin{abstract}

Recently, leveraging on the development of end-to-end convolutional neural networks (CNNs), deep stereo matching networks have achieved remarkable performance far exceeding traditional approaches. However, state-of-the-art stereo frameworks still have difficulties at finding correct correspondences in texture-less regions, detailed structures, small objects and near boundaries, which could be alleviated by geometric clues such as edge contours and corresponding constraints. To improve the quality of disparity estimates in these challenging areas, we propose an effective multi-task learning network, \emph{EdgeStereo}, composed of a disparity estimation branch and an edge detection branch, which enables end-to-end predictions of both disparity map and edge map. To effectively incorporate edge cues, we propose the edge-aware smoothness loss and edge feature embedding for inter-task interactions. It is demonstrated that based on our unified model, edge detection task and stereo matching task can promote each other. In addition, we design a compact module called residual pyramid to replace the commonly-used multi-stage cascaded structures or 3-D convolution based regularization modules in current stereo matching networks. By the time of the paper submission, \emph{EdgeStereo} achieves state-of-art performance on the FlyingThings3D dataset, KITTI 2012 and KITTI 2015 stereo benchmarks, outperforming other published stereo matching methods by a noteworthy margin. \emph{EdgeStereo} also achieves  comparable generalization performance for disparity estimation because of the incorporation of edge cues.


\end{abstract}

\section{Introduction}

Stereo matching and depth estimation from stereo images have a wide range of applications, including robotics \cite{schmid2013stereo}, medical imaging \cite{nam2012application}, remote sensing \cite{shean2016automated}, 3-D computational photography \cite{barron2015fast} and autonomous driving \cite{menze2015object}. The main goal of stereo matching is to find corresponding pixels from two viewpoints, producing dense depth data in a cost-efficient manner. Given a rectified stereo pair, supposing a pixel $(x,y)$ in the left image has a disparity $d$, its corresponding point can be found at $(x-d,y)$ in the right image. Consequently, the depth of this pixel can be obtained by $\frac{fT}{d}$, where $f$ denotes the focal length and $T$ denotes the baseline distance between two cameras.

As a classical research topic for decades, stereo matching was traditionally formulated as a multi-stage optimization problem \cite{hirschmuller2005accurate,zhang2007estimating} with a popular four-step pipeline \cite{scharstein2002taxonomy} including matching cost computation, cost aggregation, optimization and disparity refinement. For instance, the popular Semi-Global Matching (SGM) \cite{hirschmuller2005accurate} adopted dynamic programming to optimize an energy function for a locally optimal matching cost distribution, followed by several post-processing functions. However, performance of the traditional stereo matching methods is severely limited by hand-crafted matching cost descriptors, and engineered energy function and optimization procedures.

\begin{figure*}[tb]
\centering
\includegraphics[width=1\textwidth]{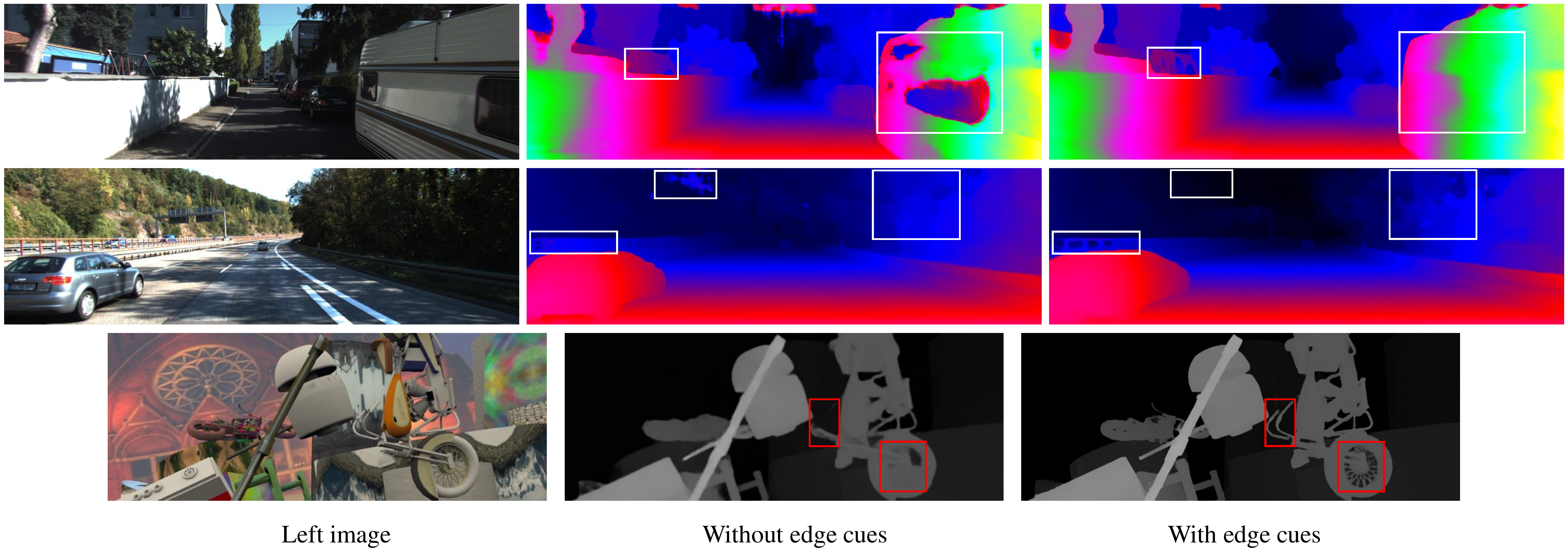}
\caption{Examples of disparity estimates. From top to bottom: KITTI 2012, KITTI 2015 and FlyingThings3D datasets. From left to right: left image, disparity map from the disparity sub-network without edge cues, and disparity map from \emph{EdgeStereo}. As shown in the boxes, disparity estimates are more accurate in ill-posed regions (reflective regions), detailed structures, and sky where ground-truth disparities don't exist, after incorporating edge cues. (Better zoom in and view in color)}
\label{intro}
\end{figure*}

Recently, with the development of convolutional neural networks, stereo matching is cast as a learning task. Early CNN-based stereo matching methods \cite{zbontar2016stereo,luo2016efficient} mainly focused on representing image patches with powerful deep features then conducting matching cost computations. Significant gains are achieved compared with the traditional approaches, however most of these stereo networks have following limitations: 1) high computational burden from multiple forward passes for all potential disparities; 2) limited receptive field and the lack of context information to infer reliable correspondences in ill-posed regions; 3) still used post-processing functions which are hand-engineered with a number of empirically set parameters. Alternatively, the end-to-end disparity estimation networks seamlessly integrate all steps in the stereo matching pipeline for joint optimization \cite{mayer2016large}, producing dense disparity maps from stereo images directly. 2-D encode-decoder structures with cascaded refinement \cite{pang2017cascade} and regularization modules composed of 3-D convolutions \cite{kendall2017end} are two of the most popular structures in current end-to-end stereo matching networks, which are demonstrated to be extremely effective for disparity estimation, benefitting from a large number of training data.

However, even the state-of-the-art end-to-end stereo networks may find it difficult to overcome local ambiguities in ill-posed regions, such as repeated patterns, textureless regions or reflective surfaces. Producing accurate disparity estimates is also challenging for detailed structures, small objects and near boundaries. Several end-to-end stereo networks \cite{kendall2017end,chang2018pyramid} enlarged their receptive fields through a stack of convolutional layers or spatial pooling modules, to encode context information for ill-posed regions where many potential correspondences exist. However, overlooking global context will inevitably lose high-frequency information that helps generate fine details in disparity maps. Moreover, without geometric constraints being utilized, most of these stereo matching networks are over-fitted, and corresponding generalization capabilities are poor.

Humans perform binocular alignment well at ill-posed regions by perceiving object boundaries in scenes, which is an important clue for detecting depth changes between background and foreground, and maintaining depth consistency for individuals. Consequently, combining semantic-meaningful edge cues provides beneficial geometric knowledge for stereo matching, meanwhile high-frequency feature representations are also supplemented for stereo networks with large receptive fields. For example, in Fig. \ref{intro}, the disparity estimates in vehicle's reflective surface, sky and image details are refined after incorporating edge cues.

This paper also asks a question, which has not been discussed by other CNN-based stereo methods yet, can stereo matching help other computer vision tasks through a unified convolutional neural network? In dense and high-quality disparity maps, the accurate depth boundary serves as an implicit but helpful geometric constraint for tasks like edge detection, semantic segmentation or instance segmentation \emph{etc.} Considering the aforementioned problems in stereo matching, we intend to explore the mutual exploitation of stereo and edge information in a unified model, which is the main contribution of this paper.

Consequently, we design an effective multi-task learning network, \emph{EdgeStereo}, that incorporates edge cues and corresponding regularization into the stereo matching pipeline. \emph{EdgeStereo} consists of a disparity estimation branch and an edge detection branch, meanwhile two sub-networks share the shallow part of the backbone and low-level features. During training, interactions between the tasks are two-folds: firstly, the edge features are embedded into the disparity branch providing fine-grained representations; secondly, the edge map in the edge branch is utilized in our proposed edge-aware smoothness loss, which guides the multi-task learning in \emph{EdgeStereo}. During testing, end-to-end predictions of both disparity map and edge map are enabled.

Basically, we first use ResNet \cite{he2016deep} to extract image descriptors from a stereo pair and compute a cost volume by means of a correlation layer \cite{Dosovitskiy2015FlowNet}. Then the concatenation of left image descriptor, matching cost volume and edge features are fed to a regularization module, to regress a full-size disparity map. In \emph{EdgeStereo}, we design an hourglass structure composed of 2-D convolutions as the regularization module. Different from the decoder in DispNet \cite{mayer2016large}, cascaded structures in \cite{pang2017cascade,liang2017learning} and 3-D convolution based regularization modules in \cite{kendall2017end,chang2018pyramid}, we propose a compact module called residual pyramid as the decoder for disparity regression. Some works use the same principle of residual pyramid for both optical flow \cite{sun2018pwc} and stereo matching \cite{tonioni2019real}. In residual pyramid, the disparities are directly estimated only at the smallest scale and the residual signals are predicted at other scales, hence making \emph{EdgeStereo} an efficient one-stage model. Compared with other regularization modules, the proposed residual pyramid reduces the amount of parameters and improve the generalization capability as well as model interpretability. Finally, the produced disparity map and edge map are both optimized under the guidance of the edge-aware smoothness loss.

In \emph{EdgeStereo}, the edge branch and disparity branch are both fully-convolutional so that end-to-end training can be conducted. Considering there is no dataset containing both edge annotations and ground-truth disparities, we propose an effective multi-stage training method. After multi-task learning, stereo matching task and edge detection task are both improved quantitatively and qualitatively.
For stereo matching, \emph{EdgeStereo} achieves the best performance on the FlyingThings3D dataset \cite{mayer2016large}, KITTI 2012 \cite{geiger2012we} and KITTI 2015 \cite{menze2015object} stereo benchmarks compared with all published stereo methods. Particularly in the evaluation of ``Reflective Regions'' in the KITTI 2012 benchmark, \emph{EdgeStereo} also outperforms other methods by a noteworthy margin. For edge detection, after the multi-task learning on a stereo matching dataset, edge predictions from \emph{EdgeStereo} are improved, even if the stereo matching dataset does not contain ground-truth edge labels for training.

Our contributions and achievements are summarized below.

-- We propose the multi-task learning network \emph{EdgeStereo} that incorporates edge detection cues into the disparity estimation pipeline. For effective multi-task interactions, we design the edge feature embedding and propose the edge-aware smoothness loss. It is demonstrated that edge detection task and stereo matching task can promote each other based on our unified model. As far as we know, EdgeStereo is the first multi-task learning framework for stereo matching and edge detection.

-- We design the residual pyramid, which is a compact decoder structure for disparity regression.

-- Our method achieves state-of-the-art performance on the FlyingThings3D dataset, KITTI 2012 and KITTI 2015 stereo benchmarks, outperforming other stereo methods by a noteworthy margin. In addition, \emph{EdgeStereo} is demonstrated with a comparable generalization capability for disparity estimation.

The rest of paper is organized as follows. After reviewing related work in Section \ref{s2}, we introduce the overall architecture, residual pyramid and edge regularization in Section \ref{s3}. Then in Section \ref{s4}, we conduct detailed ablation studies to confirm the effectiveness of our design, and we compare \emph{EdgeStereo} with other state-of-the-art stereo matching methods. Finally, Section \ref{s5} concludes the paper.

\section{Related Work}
\label{s2}
\subsection{Stereo Matching}

Following the traditional stereo pipeline \cite{scharstein2002taxonomy}, a great number of hand-engineered stereo matching methods have been proposed for matching cost computation \cite{geiger2010efficient,heise2015fast}, aggregation \cite{zhang2007estimating} and optimization \cite{hirschmuller2005accurate,kolmogorov2001computing,klaus2006segment}. Over the past few years, the convolutional neural network has been introduced to solve various problems in traditional stereo methods, and state-of-the-art performance is achieved. We hereby review stereo matching with emphasis placed on CNN-based methods, which can be roughly divided into three categories.

\subsubsection{Non-end-to-end Stereo Matching}

For non-end-to-end stereo methods, a CNN is introduced to replace one or more components in the legacy stereo pipeline. The first success of convolutional neural network for stereo matching was achieved by substituting hand-crafted matching cost with deep metrics. Zbontar and LeCun \cite{zbontar2015computing} first introduced a deep siamese network to measure the similarity between two $9\times9$ image patches. Luo \emph{et al.} \cite{luo2016efficient} accelerated matching cost calculation by an inner-product layer and proposed to learn a multi-label classification model over all possible disparities. Chen \emph{et al.} \cite{chen2015deep} proposed an embedding model fusing multi-scale features for matching cost calculation. Concurrently, Zagoruyko \emph{et al.} \cite{zagoruyko2015learning} investigated various CNN structures to compare image patches. In these methods, after obtaining a cost volume through a CNN, several non-learned post-processing functions are followed, including
the cross-based cost aggregation, semi-global matching, left-right consistency check, sub-pixel enhancement and bilateral filtering \cite{mei2011building}.

Besides the similarity measurement, deep neural networks could also be used in other sub-tasks. Gidaris \emph{et al.} \cite{gidaris2016detect} substituted hand-crafted disparity refinement functions with a three-stage network that detects, replaces and refines erroneous predictions. Shaked and Wolf \cite{shaked2016improved} introduced an network pooling global information from a cost volume for initial disparity prediction. Seki \emph{et al.} \cite{seki2017sgm} raised the SGM-Net framework that predicts SGM penalties for regularization. Knobelreiter \emph{et al.} \cite{knobelreiter2017end} learned smoothness penalties through a CRF model for energy function optimization. In these methods, a number of hand-crafted regularization functions are still necessary to achieve comparable results.

\begin{figure*}[tb]
\centering
\includegraphics[width=1.02\textwidth]{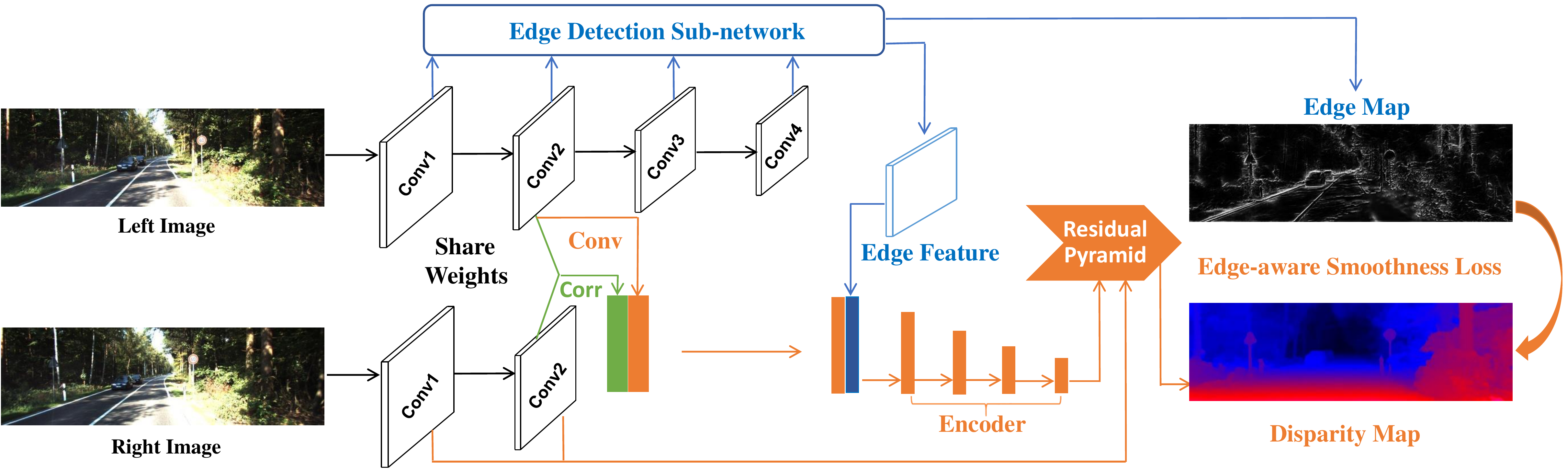}
\caption{Overall architecture. \emph{EdgeStereo} consists of a disparity estimation branch and an edge detection branch,  sharing the shallow part of the backbone. Taking a fused representation as input, the regularization module (encoder and the residual pyramid) produces a full-size disparity map. To obtain disparity maps with fine details and conduct multi-task learning, edge cues are incorporated into the disparity branch by the edge feature embedding and edge-aware smoothness loss.}
\label{archi}
\end{figure*}

\subsubsection{End-to-end Stereo Matching}

Inspired by other pixel-wise labeling tasks, fully-convolutio-\\nal networks (FCN) \cite{long2015fully} were carefully designed to regress disparity maps from stereo inputs without post-processing.  All components in the legacy stereo pipeline are combined for joint optimization. Mayer \emph{et al.} \cite{mayer2016large} proposed the first end-to-end disparity estimation network called DispNet, in which an 1-D correlation layer was proposed for the cost calculation, and an encoder-decoder structure with shor-tcut connections \cite{ronneberger2015u} was designed for disparity regression. They also released a large synthetic stereo matching dataset, making it possible to pretrain an end-to-end stereo matching network without over-fitting. Based on DispNet, Pang \emph{et al.} \cite{pang2017cascade} proposed a two-stage architecture called cascade residual learning (CRL) where the first stage gives initial predictions, and the second stage learns residuals. Liang \emph{et al.} \cite{liang2017learning} extended DispNet and designed a different disparity refinement sub-network, in which two stages are combined for joint learning based on the feature constancy.

Different from DispNet and its variants, Kendall \emph{et al.} \cite{kendall2017end} raised GC-Net in which the feature dimension is not collapsed when constructing a cost volume, and 3-D convolutions are used to regularize the cost volume and incorporate more context from the disparity dimension. Inspired by GC-Net, Chang \emph{et al.} \cite{chang2018pyramid} employed a spatial pyramid pooling module to extract multi-scale representations and designed a stacked 3-D CNN for cost volume regularization. Tulyakov \emph{et al.} designed a practical deep stereo (PDS) network \cite{tulyakov2018practical} where memory footprint is reduced by compressing concatenated image descriptors into compact representations before feeding to the 3-D regularization module.

Several end-to-end stereo methods focused on designing specific functional modules. Lu \emph{et al.} \cite{lu2018sparse} introduced an efficient stereo matching method based on their proposed sparse cost volume. Yu \emph{et al.} \cite{yu2018deep} proposed a two-stream architecture for better generation and selection of cost aggregation proposals from cost volumes. Jie \emph{et al.} \cite{jie2018left} proposed a left-right comparative recurrent (LRCR) model for stereo matching, which is the first end-to-end network that incorporates left-right consistency check into disparity regression using stacked convolutional LSTM.

Our method also enables end-to-end predictions of full-size disparity maps. Rather than regressing disparities from multi-stage cascaded structures or 3-D convolution based regularization modules, which makes stereo networks over-parameterized, \emph{EdgeStereo} is an efficient one-stage network with better model interpretability because of the proposed residual pyramid.

\subsubsection{Unsupervised Stereo Matching}

Over the past few years, based on spatial transformation and view synthesis, several unsupervised learning methods were proposed for stereo matching without the need of ground-truth disparities during training. Tonioni \emph{et al.} \cite{tonioni2017unsupervised} propose a novel unsupervised adaptation approach that enables to finetune a deep learning stereo model without any ground-truth. Based on GC-Net, Zhong \emph{et al.} \cite{zhong2017self} proposed an unsupervised self-adaptive stereo matching network under the guidance of photometric errors. Zhou \emph{et al.} \cite{zhou2017unsupervised} presented an unsupervised framework to learn matching costs iteratively, supervised by a left-right consistency loss.

The view-synthesis based unsupervised stereo matching can also be adapted to unsupervised monocular depth estimation. Garg \emph{et al.} \cite{garg2016unsupervised} proposed the first unsupervised network for single-view depth estimation, in which per-pixel disparity is learned driven by an image reconstruction loss in a calibrated stereo environment. Based on \cite{garg2016unsupervised}, Godard \emph{et al.} \cite{godard2017unsupervised} developed a fully-differentiable structure for photometric error minimization and designed a left-right disparity consistency term for regularization. Kuznietsov \emph{et al.} \cite{kuznietsov2017semi} proposed a semi-supervised approach where ground-truth depths and unsupervised binocular alignment losses are both used to train the monocular depth estimation network.

Although unsupervised methods get rid of the dependence on ground truth disparities, they are still not comparable with supervised methods (Our \emph{EdgeStereo}). In addition, in unsupervised stereo networks, various smoothness terms were designed to regularize disparity maps. However, these image gradient based regularization terms are not robust and are inferior to our proposed edge-aware smoothness loss, as demonstrated in Section \ref{s4}.

\subsection{Edge Detection}

Edge detection is a classical low-level vision task. Early methods focused on designing hand-engineered descriptors using low-level cues such as intensity and gradient \cite{canny1986computational,dollar2006supervised}, then employing sophisticated learning paradigm for classification \cite{dollar2015fast}. Recently, inspired by FCN \cite{long2015fully}, Xie \emph{et al.} \cite{xie2015holistically} designed the first end-to-end edge detection network called holistically-nested edge detector (HED), in which all output layers in side branches are connected to aggregate coarse-to-fine predictions for a final output. Liu \emph{et al.} \cite{liu2016learning} achieved some improvements compared with HED by employing relaxed labels to guide the training process. Liu \emph{et al.} \cite{liu2017richer} modified the network structure of HED, combining all meaningful convolutional features in the backbone for prediction. Since the structures of these fully-convolutional edge detection networks are concise and efficient, we propose a similar edge detection sub-network in our unified model, providing semantic-meaningful edge features and geometric constraints for stereo matching.

\subsection{Deep Multi-task Network}

Kendall \emph{et al.} \cite{kendall2018multi} propose a multi-task learning model learning per-pixel depth regression, semantic and instance segmentation from a monocular input image. Ramirez \emph{et al.} \cite{ramirez2018geometry} propose a semi-supervised framework aimed at joint semantic segmentation and depth estimation. Cheng \emph{et al.} \cite{cheng2017segflow} proposed an end-to-end architecture called SegFlow, which enables the joint learning of video object segmentation and optical flow. This model consists of a FCN \cite{long2015fully} based segmentation branch and a FlowNet \cite{Dosovitskiy2015FlowNet} based flow branch, in which feature maps of two tasks are concatenated and two branches are trained alternately. However, this multi-task architecture requires the dataset containing all types of ground-truth labels for different tasks during training, which limits its adaptation ability to other tasks. Mao \emph{et al.} \cite{mao2017can} also proposed a multi-task network called HyperLearner to help pedestrian detection. Diverse features from different tasks, such as semantic segmentation and edge detection \emph{etc}, provide various representations which are concatenated with the backbone pedestrian detection network. However, multi-task learning is not conducted in this architecture because losses in the detection branch could not be propagated back to other branches, meanwhile geometric knowledge from other tasks is not fully exploited. With a similar motivation to ours, Yang \emph{et al.} \cite{yang_segstereo} proposed a unified model called SegStereo for semantic segmentation and disparity estimation, in which semantic features are utilized and a semantic softmax loss is introduced to improve
the quality of disparity estimates especially in texture-less regions. However, in SegStereo, disparity estimation could not help  semantic segmentation, and incorporating high-level semantic cues causes the loss of high-frequency information.

In this paper, we incorporate edge cues into the disparity estimation pipeline obtaining a unified multi-task learning architecture. The embedded edge features provide semantic-meaningful and high-frequency representations, meanwhile the proposed edge-aware smoothness loss carries beneficial geometric constraints for effective multi-task learning. In addition, the training dataset is not required to contain all types of ground-truth labels for different tasks. Comparisons with our previous work \cite{song2018stereo} can be found in Section 4.7. Our latest \emph{EdgeStereo} model achieves state-of-the-art performance on the FlyingThings3D, KITTI 2012 and KITTI 2015 datasets, outperforming all published stereo matching methods.

\section{Approach}
\label{s3}

We present \emph{EdgeStereo}, which is an effective multi-task learning network for stereo matching and edge detection. We would like to learn an optimal mapping from a stereo pair to a disparity map and an edge map corresponding to the reference image. However, we do not intend to design a machine learning model acting as a complete black box. Hence we develop several differentiable modules representing major components in the stereo matching pipeline, meanwhile leveraging geometric knowledge from edge detection.

In this section, we first present the basic network architecture of \emph{EdgeStereo}. Then we introduce a key module for disparity regression: residual pyramid. Next we detail the incorporation strategies of edge cues, including the edge feature embedding and  edge-aware smoothness loss. Furthermore we show how to conduct multi-task learning through the proposed multi-stage training method. Finally we introduce the detailed structure of \emph{EdgeStereo} and give a layer-by-layer definition.

\subsection{Basic Network Architecture}

The overall architecture of \emph{EdgeStereo} is shown in Fig. \ref{archi}, in which two branches share the shallow part of the backbone. For the disparity branch, instead of using multiple networks for different components, we combine all modules into a single network composed of extraction, matching and regularization modules. In addition, we propose the residual pyramid so that disparity refinement is not required, making \emph{EdgeStereo} an efficient one-stage architecture.

The extraction module outputs the image descriptors $\mathbb{F}^l$ and $\mathbb{F}^r$ carrying local semantic features for left and right images through the shared backbone. Then the matching module performs an explicit correlation operation using the 1-D correlation layer from \cite{Dosovitskiy2015FlowNet}, capturing coarse correspondences between two descriptors for each potential disparity. Then the cost volume $\mathbb{F}_c$ is obtained. To preserve detailed information in the left image descriptor, we apply a convolutional block on $\mathbb{F}^l$ obtaining the refined representation $\mathbb{F}_r^l$.

The edge branch shares shallow layer representations with the disparity branch, meanwhile remaining convolutional features in the backbone are used to produce edge features and a semantic-meaningful edge map of the left image. Similarly, we also apply a convolutional block and obtain the transformed edge features $\mathbb{F}_e^l$. The left image descriptor $\mathbb{F}_r^l$,  cost volume $\mathbb{F}_c$ and transformed edge features $\mathbb{F}_e^l$ are concatenated as the hybrid feature representation $\mathbb{F}_h$, which is referred to as \emph{edge feature embedding}.

The regularization module takes $\mathbb{F}_h$ as input, regularizes it and produces a dense disparity map. This module is usually implemented as an hourglass structure with shortcut connections between encoder and decoder. We design a 2-D convolution based encoder different from existing encoder structures \cite{mayer2016large,pang2017cascade,liang2017learning}, in which the residual block in ResNet \cite{he2016deep} is adopted as the basic block for better information flow. Furthermore, we design the residual pyramid as the decoder, where shortcut connections are not required and a coarse-to-fine learning mechanism is employed. In the residual pyramid, which takes the sub-sampled hybrid feature representation as input, disparities are directly regressed only at the smallest scale and residual signals are predicted for refinement at other scales. The smallest scale of disparity estimates is not fixed, which can be specified by applying several deconvolutional blocks on the encoder. Finally, disparity and residual learning are regularized across multiple scales in the residual pyramid, guided by the \emph{edge-aware smoothness loss} .

\subsection{Residual Pyramid}

For disparity refinement, existing methods \cite{gidaris2016detect,pang2017cascade,liang2017learning} cascaded additional networks on the initial disparity prediction network, learning multi-scale residual signals. However, these cascaded structures that model the joint space of multiple networks are over-parameterized. In addition, the initial full-size disparity estimates are quite accurate in ordinary areas, but residual learning without geometric constraints is not easy in ill-posed regions, leading to approximately zero residual signals.

\begin{figure}[tb]
\centering
\includegraphics[width=8.5cm]{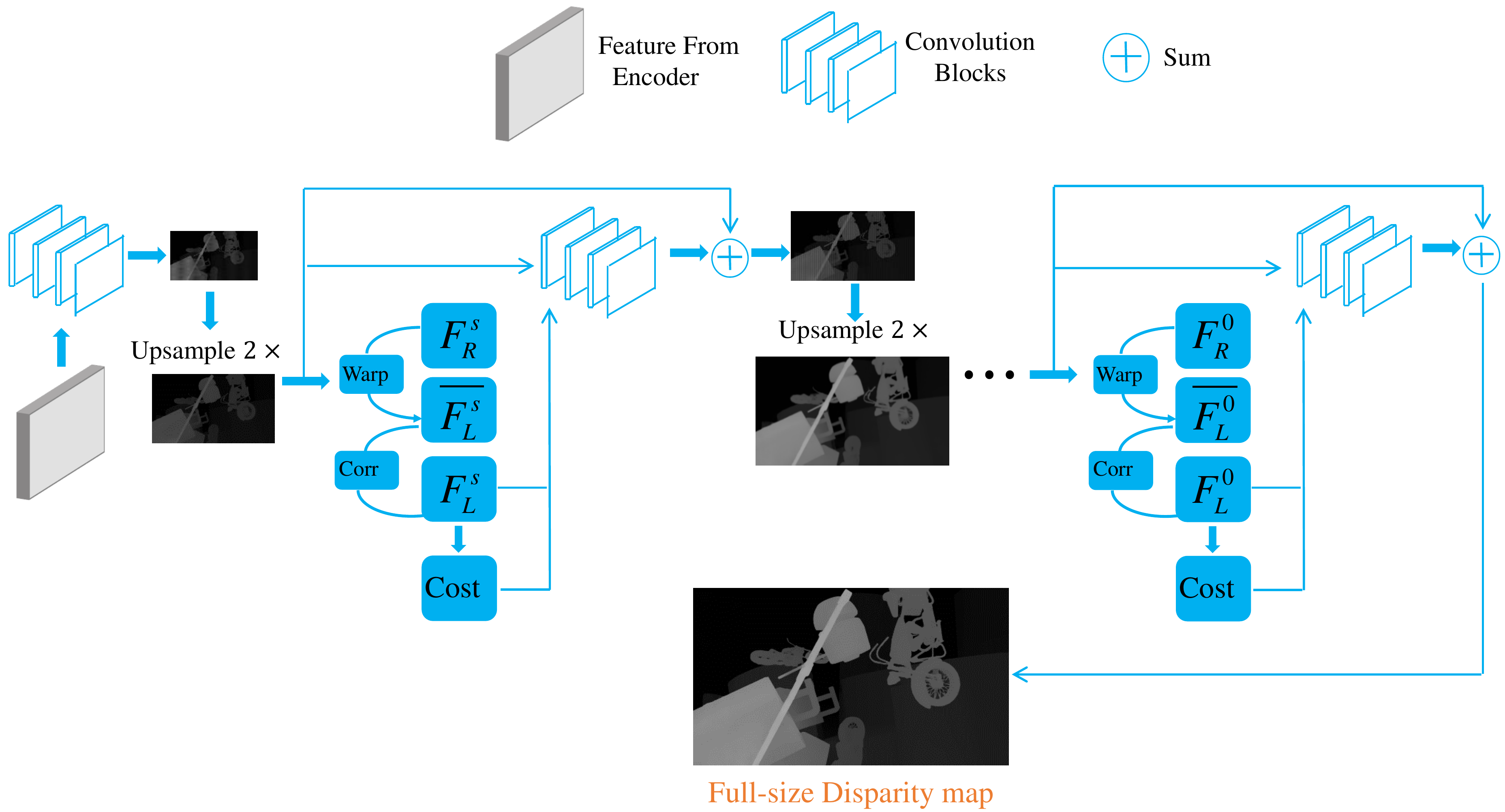}
\caption{Overview of the residual pyramid. Disparities are directly estimated only at the smallest scale and residual signals are produced at other scales for refinement, making \emph{EdgeStereo} an efficient one-stage architecture. Convolutional blocks are adopted to produce initial disparities and residual signals.}
\label{res}
\end{figure}

\begin{figure*}[tb]
\centering
\includegraphics[width=0.95\textwidth]{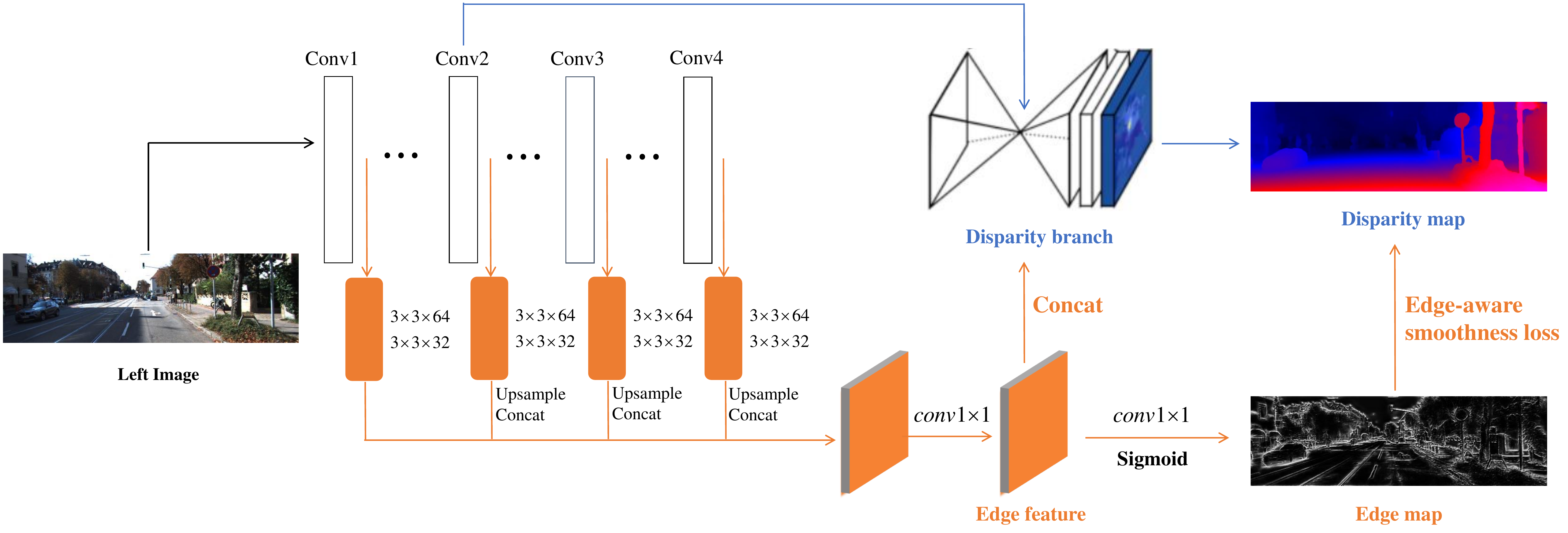}
\caption{Overview of the edge branch. To preserve subtle details, fine-grained edge features are supplemented, meanwhile the edge-aware smoothness loss imposes beneficial object-level constraints on disparity training and multi-task learning.}
\label{edge_branch}
\vspace{-1mm}
\end{figure*}

To alleviate the aforementioned problems, we adopt the coarse-to-fine residual learning mechanism, and propose the residual pyramid that enables learning and refining disparities in a single decoder. In general, estimating disparities at the smallest scale is easier since the searching range is narrow, meanwhile significant areas are emphasized and less details are required. Therefore the initial disparity estimate acts as a good starting point. To obtain an up-sampled disparity map with extra details, we continuously predict residual signals at other scales, utilizing high-frequency representations and beneficial geometric constraints from edge cues. In addition, the coarse-to-fine residual learning mechanism benefits the overall training and alleviates the problem of over-fitting.

The structure of the residual pyramid is flexible. The number of scales $S$ in the residual pyramid is dependent on the total sub-sampling factor of the input volume. The disparity map at the smallest scale is denoted as $d_{S-1}$ ($\frac{1}{2^{S-1}}$ of the full size), then it is continuously up-sampled and refined with the residual signal
 $r_s$ at scale $s$, until the full-size disparity map $d_0$ is obtained. As shown in Eq. (\ref{sum}), $u(\cdot)$ denotes up-sampling by a factor of $2$ and $s$ denotes the scale.

\begin{equation}
\\ \\ d_s=u(d_{s+1})+r_s, \, 0\le s< S\
\label{sum}
\end{equation}

Existing encoder-decoder structures \cite{pang2017cascade,liang2017learning,chang2018pyramid,kendall2017end} rely on shortcut connections to fuse high-level and fine-grained representations, which complicates the feature space of disparity regularization and makes stereo networks less explainable. Conversely we leverage the knowledge from stereo geometry and geometric constraints from edge cues to learn residuals, without the need for shortcut connections for direct disparity regression. As shown in Fig. \ref{res}, to predict the residual signal at scale $s$, we first use the up-sampled disparity map $u(d_{s+1})$ to warp the right image features $F_{R}^{s}$ at scale $s$ in the backbone, obtaining the synthesised left image features $\overline{F_{L}^{s}}$. Then we perform 1-D correlation between the synthesised and real left image features $F_{L}^{s}$, obtaining a cost volume $cost$ that characterizes left-right feature consistency based on  existing disparity estimates $u(d_{s+1})$. We use feature representations rather than raw pixel intensities for binocular warping, because they are more robust to photometric appearances and can encode local context information. Finally, the concatenation of the left image features $F_{L}^{s}$, up-sampled disparity map $u(d_{s+1})$ and cost volume $cost$ are processed by several convolutional blocks, producing the residual signal $r_s$. At each scale, residuals are explicitly supervised by the edge-aware smoothness loss and the difference between disparity estimates and ground-truth labels, learning sharp transitions at object boundaries.

\subsection{Incorporation of Edge Cues}

As shown in Fig. \ref{intro}, the disparity estimation sub-network without edge cues works well in ordinary areas, where matching clues can be easily captured. However, accurate predictions for detailed structures are not given due to the lack of fine-grained representations in the deep stereo network. In addition, disparity estimates in reflective regions and near boundaries are not accurate due to the lack of geometric knowledge and constraints. Hence we incorporate edge cues to regularize the disparity learning, as shown in Fig. \ref{edge_branch}.

The first incorporation of edge cues is the edge feature embedding. After the extraction module and matching module, the transformed edge features $\mathbb{F}_e^l$ are concatenated with the left image descriptor $\mathbb{F}_r^l$ and cost volume $\mathbb{F}_c$ before fed to the regularization module. Advantages are: 1) The cooperated edge branch shares the efficient computation and effective representations with the disparity branch; 2) The edge features $\mathbb{F}_e^l$  supplement fine-grained representations from image details and boundaries, bringing beneficial scene priors to the regularization module; 3) The aggregation of feature maps from different tasks facilitates the multi-task interactions during training.

The second incorporation of edge cues is the proposed edge-aware smoothness loss, denoted as $L_{sm}$. We encourage smooth disparity estimates in local neighborhoods and the loss term penalizes drastic depth changes in flat regions. To allow for depth discontinuities at object boundaries, previous methods \cite{Godard2016Unsupervised,zhong2017self} weight smoothness regularization terms based on the image gradient which is not robust to various photometric appearances. Differently, we raise the edge-aware smoothness loss based on the edge map gradient from the edge detection sub-network, which is more semantic-meaningful than the variation of raw pixel intensities. As shown in Eq. (\ref{smoo}), where $N$ denotes the number of valid pixels, $\partial d$ denotes the disparity gradient, $\partial\mathcal{E}$ denotes the edge map gradient and the hyper-parameter $\beta$ controls the intensity of this smoothness regularization term.

\begin{equation}
\\ \\ L_{sm}=\frac{1}{N}\sum_{i,j}|\partial_xd_{i,j}|e^{-\beta|\partial_{x}\mathcal{E}_{i,j}|}+|\partial_yd_{i,j}|e^{-\beta|\partial_{y}\mathcal{E}_{i,j}|}
\label{smoo}
\end{equation}

The edge-aware smoothness loss also facilitates the multi-task learning in \emph{EdgeStereo}. During training, the edge-aware smoothness loss is propagated back to the disparity branch and edge branch. The disparity estimates and edge predictions are improved simultaneously, until sharp disparities and fine edge predictions are obtained. The regularization term utilizes the geometric prior that depth boundaries in a disparity map should be consistent with edge contours in the scene, which imposes beneficial object-level constraints on disparity training and multi-task learning.

The experiments in Section \ref{s4} demonstrate that incorporating edge cues can effectively help disparity estimation in detailed structures, reflective regions and near boundaries. In addition, edge predictions are also improved after the multi-task learning on a stereo matching dataset, even if edge annotations are not provided for training.

\begin{table}[tb]
\scriptsize
\renewcommand{\captionfont}{\scriptsize}
\centering
\caption{Layers in \emph{EdgeStereo\_RP\_2}. \textbf{K} means kernel size, \textbf{S} means stride, \textbf{I/O} means input/ouput channels. ``,'' means concat and ``+'' means element-wise summation.}
\vspace{-1.5mm}
\label{t1}
\begin{tabularx}{9.8cm}{p{0.9cm}|p{1.9cm}|p{0.05cm}p{0.05cm}p{0.9cm}<{\centering}|p{2.4cm}|X<{\centering}}
\toprule
Type & Layer & K & S & c I/O  & Input & Resolution\\
\midrule
\multicolumn{7}{c}{\emph{1. Extraction Module}} \\
\midrule
\multirow{2}{*}{Conv} & {conv1\_1a}  & \multirow{2}{*}{3} & \multirow{2}{*}{2} & \multirow{2}{*}{3/64} & left image & 1/2 \\
& {conv1\_1b} & & & & right image & \\
\multirow{2}{*}{Conv} & {conv1\_2a}  & \multirow{2}{*}{3} & \multirow{2}{*}{1} & \multirow{2}{*}{64/64} & conv1\_1a & 1/2 \\
& {conv1\_2b} & & & & conv1\_1b & \\
\multirow{2}{*}{Conv} & {conv1\_3a}  & \multirow{2}{*}{3} & \multirow{2}{*}{1} & \multirow{2}{*}{64/128} & conv1\_2a & 1/2 \\
& {conv1\_3b} & & & & conv1\_2b & \\
\midrule
\multicolumn{7}{c}{\emph{2. Edge Detection Sub-network}} \\
\midrule
Pooling & max\_pool1 & 3 & 2 & 128/128 & conv1\_3a & 1/4\\
ResNet50 & conv2\_1\ding{213}conv2\_3  & 3 & 1 & 128/256 & max\_pool1 & 1/4\\
ResNet50 & conv3\_1\ding{213}conv3\_4  & 3 & 2 & 256/512 & conv2\_3 & 1/8\\
ResNet50 & conv4\_1\ding{213}conv4\_6  & 3 & 1 & 512/1024 & conv3\_4 & 1/8\\
Conv & conv1\_3\_edge\_1 & 3 & 1 & 128/64 & conv1\_3a & 1/2\\
Conv & conv1\_3\_edge & 3 & 1 & 64/32 & conv1\_3\_edge\_1 & 1/2\\
Conv & conv2\_3\_edge\_1 & 3 & 1 & 256/64 & conv2\_3 & 1/4\\
Conv & conv2\_3\_edge & 3 & 1 & 64/32 & conv2\_3\_edge\_1 & 1/4\\
Interp & conv2\_3\_edge\_i & \verb|-| & 2 & \verb|-| & conv2\_3\_edge & 1/2\\
Conv & conv3\_4\_edge\_1 & 3 & 1 & 512/64 & conv3\_4 & 1/8\\
Conv & conv3\_4\_edge & 3 & 1 & 64/32 & conv3\_4\_edge\_1 & 1/8\\
Interp & conv3\_4\_edge\_i & \verb|-|  & 4 & \verb|-| & conv3\_4\_edge & 1/2\\
Conv & conv4\_6\_edge\_1 & 3 & 1 & 1024/64 & conv4\_6 & 1/8\\
Conv & conv4\_6\_edge & 3 & 1 & 64/32 & conv4\_6\_edge\_1 & 1/8\\
Interp & conv4\_6\_edge\_i & \verb|-|  & 4 & \verb|-| & conv4\_6\_edge & 1/2\\
\multirow{2}{*}{Conv} & \multirow{2}{*}{conv\_edge} & \multirow{2}{*}{1} & \multirow{2}{*}{1} & \multirow{2}{*}{128/128} & {\tiny conv1\_3\_edge,conv2\_3\_edge\_i,} &  1/2\\
& & & & & {\tiny conv3\_4\_edge\_i,conv4\_6\_edge\_i} & \\
Conv & {\tiny edge\_score (no BN/ReLU)} & 1 & 1 & 128/1 & conv\_edge & 1/2\\
Sigmoid & edge\_map & \verb|-| & \verb|-| & \verb|-| & edge\_score  & 1/2\\
\midrule
\multicolumn{7}{c}{\emph{3. Matching Module and Edge Feature Embedding}} \\
\midrule
\multirow{2}{*}{Conv} & {conv\_trans\_a}  & \multirow{2}{*}{3} & \multirow{2}{*}{1} & \multirow{2}{*}{128/128} & conv1\_3a & 1/2\\
& {conv\_trans\_b} & & & & conv1\_3b & \\
\multirow{2}{*}{Corr} & \multirow{2}{*}{Corr\_1d}  & \multirow{2}{*}{\verb|-|} & \multirow{2}{*}{\verb|-|} & \multirow{2}{*}{128/97} & conv\_trans\_a & 1/2\\
&  & & & & conv\_trans\_b & \\
Pooling & pool\_corr & 3 & 2 & 225/225 & conv\_trans\_a,Corr\_1d & 1/4\\
Conv &conv\_edge\_trans & 3 & 2 & 128/64 & conv\_edge & 1/4\\
Concat & hybrid\_feature & \verb|-| & \verb|-| & /289 & pool\_corr,conv\_edge\_trans & 1/4\\
\midrule
\multicolumn{7}{c}{\emph{4. Encoder}} \\
\midrule
ResBlock & res2\_1\ding{213}res2\_3  & 3 & 1 & 289/256 & hybrid\_feature & 1/4\\
ResBlock & res3\_1\ding{213}res3\_4  & 3 & 2 & 256/512 & res2\_3 & 1/8\\
ResBlock & res4\_1\ding{213}res4\_6  & 3 & 1 & 512/1024 & res3\_4 & 1/8\\
ResBlock & res5\_1\ding{213}res5\_3  & 3 & 1 & 1024/2048 & res4\_6 & 1/8\\
Conv & disp\_conv5\_4 & 3 & 1 & 2048/512 & res5\_3 & 1/8\\
\midrule
\multicolumn{7}{c}{\emph{5. Decoder and Residual Pyramid}} \\
\midrule
Deconv & disp\_deconv1 & 3 & 2 & 512/256 & disp\_conv5\_4 & 1/4\\
Deconv & disp\_deconv2 & 3 & 2 & 256/128 & disp\_deconv1 & 1/2\\
Conv & disp\_conv6 & 3 & 1 & 128/32 & disp\_deconv2 & 1/2\\
Conv & {\tiny disp\_2 (no BN/ReLU)}   & 3 & 1 & 32/1 & disp\_conv6 & 1/2\\
\multirow{2}{*}{Deconv} & {disp\_ref\_a}  & \multirow{2}{*}{3} & \multirow{2}{*}{2} & \multirow{2}{*}{128/32} & conv1\_3a & 1\\
& {disp\_ref\_b} & & & & conv1\_3b & \\
Interp & up\_disp\_2 & \verb|-| & 2 & \verb|-| & disp\_2 & 1\\
\multirow{2}{*}{Warp} & \multirow{2}{*}{w\_disp\_ref\_a}  & \multirow{2}{*}{\verb|-|} & \multirow{2}{*}{\verb|-|} & \multirow{2}{*}{32/32} & disp\_ref\_b & 1\\
&  & & & & up\_disp\_2  & \\
\multirow{2}{*}{Corr} & \multirow{2}{*}{Corr\_1d\_res}  & \multirow{2}{*}{\verb|-|} & \multirow{2}{*}{\verb|-|} & \multirow{2}{*}{32/21} & disp\_ref\_a & 1\\
&  & & & & w\_disp\_ref\_a & \\
Conv & disp\_res\_conv1\_1 & 1 & 1 & 54/64 & {\tiny Corr\_1d\_res,up\_disp\_2,disp\_ref\_a} & 1\\
Conv & disp\_res\_conv1\_2 & 3 & 1 & 64/64 & disp\_res\_conv1\_1 & 1\\
Conv & disp\_res\_conv1\_3 & 3 & 1 & 64/32 & disp\_res\_conv1\_2 & 1\\
Conv & {\tiny disp\_res\_1 (no BN/ReLU)} & 3 & 1 & 32/1 & disp\_res\_conv1\_3 & 1\\
Sum & disp\_1 & \verb|-| & \verb|-| & \verb|-| & up\_disp\_2+disp\_res\_1 & 1\\
\bottomrule
\end{tabularx}
\vspace{-7mm}
\end{table}

\subsection{Multi-stage Training Method and Objective Function}

Considering there is no dataset containing both ground-truth disparities and edge annotations, in order to conduct effective multi-task learning in \emph{EdgeStereo} , we propose a multi-stage training method where training is split into three stages. Weights of the shallow part of the backbone that two tasks share are fixed in all three stages.

In the first stage, the edge detection branch is trained on an edge detection dataset, guided by the class-balanced per-pixel binary cross-entropy loss proposed in \cite{liu2017richer}.

In the second stage, we supervise disparities across $S$ scales in the residual pyramid on a stereo matching dataset. Deep supervision is adopted, forming the total loss as $C=\sum_{s=0}^{S-1}C_s$ where $C_s$ denotes the loss at scale $s$. Besides the edge-aware smoothness loss, we adopt the $L1$-loss $L_{r}$ as the disparity regression loss for supervised learning, as shown in Eq. (\ref{re}).

\begin{equation}
\\ \\ L_{r}=\frac{1}{N}{||\,d-\hat{d}\,||}_1\,,
\label{re}
\end{equation}
where $\hat{d}$ denotes the ground truth disparity. Hence the loss at scale $s$ becomes $C_s=\lambda_{r}^{s}L_{r}^{s}+\lambda_{sm}^{s}L_{sm}^{s}$, where $\lambda_{r}^{s}$ and $\lambda_{sm}^{s}$ are the loss weights for the edge-aware smoothness loss and disparity regression loss at scale $s$ respectively. In addition, weights in the edge branch are fixed in the second stage.

In the third stage, all layers in \emph{EdgeStereo} except the shared backbone are optimized on the same stereo matching dataset used in the second stage. We adopt the same deep supervision strategy as stage two and conduct effective multi-task learning.

\subsection{Model Specifications}

The backbone network is ResNet-50 \cite{he2016deep}. The shallow part that two tasks share is conv1\_1 to conv1\_3 in the ResNet-50 backbone. Hence the extracted unary features $\mathbb{F}^l$ and $\mathbb{F}^r$ are of $1/2$ spatial size to the raw
image. In the matching module, the max displacement in the 1-D correlation layer (unidirectional, leftward) is set to $96$ hence the channel number of the cost volume $\mathbb{F}_c$ is $97$.

For the edge detection branch in \emph{EdgeStereo}, we design a fully-convolutional sub-network similar to HED \cite{xie2015holistically}, while the semantic-meaningful edge features are easier to obtain in our architecture. As shown in  Fig. \ref{edge_branch}, we introduce four side branches, concatenate and fuse feature representations in all side branches, obtaining the edge features for embedding. Through another $1\times1$ convolutional layer and a sigmoid layer, an edge map is produced, in which the edge probability is given for each pixel. The edge branch uses the ResNet-50 backbone from conv1\_1 to conv4\_6 and four side branches start from conv1\_3, conv2\_3, conv3\_4 and conv4\_6 respectively. In addition, each side branch consists of two $3\times3$ convolutional blocks and a bilinear interpolation layer.

Taking the hybrid feature representation $\mathbb{F}_h$ as input, the disparity encoder contains $16$ residual blocks, similar to the structure of ResNet-50 from conv2\_3 to conv5\_3. Several convolutional layers in residual blocks are replaced with
dilated convolutional layers \cite{zhao2017pyramid} to integrate wider context information, hence the sub-sampled hybrid feature representation $\mathbb{F}_d$ from the encoder is of $1/8$ size.

As mentioned above, the structure of the residual pyramid is flexible. Depending on the smallest scale of disparity estimates, there are three different residual pyramids, denoted as $RP\_2$, $RP\_4$ and $RP\_8$ respectively: for $RP\_2$, two $3\times3$ deconvolutional blocks with stride two are applied on $\mathbb{F}_d$, hence the initial disparity map is of $1/2$ size and only full-size residual signals are required; for $RP\_4$, one $3\times3$ deconvolutional block with stride two is applied; for $RP\_8$, no deconvolutional block is applied hence the initial disparity map is of $1/8$ size. At each scale in the residual pyramid, three $3\times3$ convolutional blocks and a convolutional layer with a single output are adopted to regress disparities or residual signals.

Finally, we present a detailed layer-by-layer definition of \emph{EdgeStereo\_RP\_2}, which denotes the \emph{EdgeStereo} model with the residual pyramid $RP\_2$. As shown in Table \ref{t1}, each convolutional or deconvolutional block contains
a convolutional or deconvolutional layer, a batch normalization layer and a ReLU layer, ``Interp'' denotes the bilinear interpolation layer, ``ResNet50'' denotes a part of the ResNet-50 backbone, and ``ResBlock'' denotes the residual blocks in the encoder.

\section{Experiments}
\label{s4}

The experimental settings and results are provided. We first conduct detailed ablation studies to verify our design choices in \emph{EdgeStereo}, meanwhile we also demonstrate that stereo matching task and edge detection task can promote each other based on our unified model. Then we compare \emph{EdgeStereo} with other state-of-the-art stereo matching methods on the FlyingThings3D dataset \cite{mayer2016large}, KITTI 2012 \cite{geiger2012we} and KITTI 2015 \cite{menze2015object} stereo benchmarks. Finally, we prove that our \emph{EdgeStereo} has a comparable generalization capability for disparity estimation because of the incorporation of edge cues.

\subsection{Datasets and Evaluation Metrics}

\subsubsection{Datasets} Five publicly available stereo matching datasets are adopted for training and testing in \emph{EdgeStereo}.

(\romannumeral1) FlyingThings3D \cite{mayer2016large}: a large-scale synthetic dataset with dense ground-truth disparities, containing $22390$ stereo pairs for training and $4370$ pairs for testing. This virtual dataset contains some unreasonably large disparities; hence two specific testing protocols are widely used: \textbf{Protocol 1}, following CRL \cite{pang2017cascade}, if more than $25\%$ of disparity
values in the ground-truth disparity map are greater than $300$, the corresponding stereo pair is removed; \textbf{Protocol 2}, following PSMNet \cite{chang2018pyramid}, we only calculate errors for pixels whose ground-truth disparity $<192$. We adopt both protocols for fair comparison with other state-of-the-art stereo matching methods.

(\romannumeral2) KITTI2012 \cite{geiger2012we}: a real-world dataset with still street views from a driving car. It contains $194$ stereo pairs for training with sparse ground-truth disparities and $195$ testing pairs without ground-truth. We further divide the whole training
data into a training set ($160$ pairs) and a validation set ($34$ pairs) \footnote{The validation image indexes are 3, 15, 33, 34, 36, 45, 59, 60, 69, 71, 72, 80, 85, 88, 104, 108, 115, 146, 149, 150, 159, 161, 162, 163, 170, 172, 173, 175, 178, 179, 181, 185, 187, 188.}.

(\romannumeral3) KITTI2015 \cite{menze2015object}: a real-world dataset with dynamic street views. It contains $200$ training pairs and $200$ testing pairs. We divide the whole training data into a training set ($160$ pairs) and a validation set ($40$ pairs), following \cite{luo2016efficient}.

(\romannumeral4) CityScapes \cite{cordts2016cityscapes}: an urban scene understanding dataset. This dataset provides
$19997$ rectified stereo pairs and their disparity maps pre-computed
by the SGM algorithm \cite{hirschmuller2005accurate} in the extra training set. As illustrated in \cite{yang2018srcdisp}, combining synthetic and realistic data for pretraining is helpful. Hence we fuse the data in the extra training set of CityScapes with the training data in FlyingThings3D for \emph{EdgeStereo} pretraining.

(\romannumeral5) Middlebury 2014 \cite{scharstein2014high}: a small in-door dataset containing $15$ training pairs with dense ground-truth disparities and $15$ test pairs.

Two publicly available edge detection datasets are adopted for training and testing in \emph{EdgeStereo}.

(\romannumeral1) Multicue \cite{mely2016systematic}: a boundary and edge detection dataset with challenging natural scenes. All $100$ images are used for training. We mix the augmentation data of Multicue with the PASCAL VOC Context dataset \cite{mottaghi2014role}, to pretrain the edge detection sub-network in \emph{EdgeStereo}.

(\romannumeral2) BSDS500 \cite{arbelaez2011contour}:  a widely used edge detection dataset composed of $200$ training, $100$ validation and $200$ testing images. We only use this dataset for testing.

\subsubsection{Metrics}

For stereo matching evaluation, we adopt the end-point-error (EPE) which measures the average Euclidean distance between ground-truth and disparity estimate. We also calculate the percentage of pixels whose EPE is larger than $t$ pixels, denoted as $t$-pixel error ($>t\,px$).

For edge detection evaluation, we adopt two widely used metrics: F-measure ($\frac{2*Precision*Recall}{Precision+Recall}$) of optimal dataset scale (ODS), and F-measure of optimal image scale (OIS).

\begin{table*}[tb]
\scriptsize
\renewcommand{\captionfont}{\scriptsize}
\centering
\caption{Ablation study of edge cues and the comparison of different smoothness losses for stereo matching. For evaluation, we compute $3$-pixel error (\%) and EPE on the FlyingThings3D test set (disparity $<192$), KITTI 2012 and 2015 validation sets.}
\label{t2}
\begin{tabularx}{0.78\textwidth}{p{6cm}|X<{\centering}X<{\centering}|X<{\centering}X<{\centering}|X<{\centering}X<{\centering}}
\hlinew{0.75pt}
\multirow{2}{*}{Model} &\multicolumn{2}{c|}{FlyingThings3D} & \multicolumn{2}{c|}{KITTI 2012 val} & \multicolumn{2}{c}{KITTI 2015 val} \\
\cline{2-7}
 & $>3\,px$  & EPE & $>3\,px$  & EPE & $>3\,px$  & EPE \\
\hline
\multicolumn{7}{c}{\emph{1. Basic ablation study}} \\
\hline
Baseline model (without edge cues) & 4.443 & 0.840 & 2.844 & 0.606 & 3.192 & 0.770 \\
Baseline with edge-aware smoothness loss & 4.103 & 0.788 & 2.555 & 0.568 & 3.011 & 0.754 \\
Baseline with edge feature and edge-aware smoothness loss  & \textbf{3.940} & \textbf{0.751} & \textbf{2.385} & \textbf{0.555} & \textbf{2.839} & \textbf{0.729}\\
\hline
\multicolumn{7}{c}{\emph{2. Comparison of smoothness loss regularization}} \\
\hline
\emph{EdgeStereo} with Charbonnier smoothness loss \cite{yang_segstereo} & 3.971 & 0.756 & 2.852 & 0.691 & 3.024 & 0.769 \\
\emph{EdgeStereo} with second-order gradient smoothness loss \cite{zhong2017self} & 4.074 & 0.790 & 2.816 & 0.646 & 3.158 & 0.783 \\
\emph{EdgeStereo} (edge-aware smoothness loss) & \textbf{3.940} & \textbf{0.751} & \textbf{2.385} & \textbf{0.555} & \textbf{2.839} & \textbf{0.729}\\
\hlinew{0.75pt}
\end{tabularx}
\end{table*}

\subsection{Implementation Details}

We implement \emph{EdgeStereo} based on Caffe \cite{jia2014caffe}. The ResNet-50 model pre-trained on ImageNet \cite{deng2009imagenet} is adopted to initialize our network. The training is conducted on eight Nvidia GTX 1080Ti GPUs.

In the first training stage, we fuse the Multicue dataset with PASCAL VOC Context dataset to pretrain the edge branch. We adopt the stochastic gradient descent (SGD) with a minibatch of $16$ images. The initial learning rate is set to $0.01$ and divided by $10$ every $10K$ iterations. We set the momentum to $0.9$ and weight decay to $0.0002$. We run SGD for $40K$ iterations totally in the first stage. For data augmentation, we rotate the images in Multicue to $4$ different angles ($0$, $90$, $180$ and $270$ degrees) and flip them at each angle, and we also flip each image in the PASCAL dataset. Finally we randomly crop $513\times513$ patches for training.

In the second training stage, we fix the edge branch and pretrain \emph{EdgeStereo} on FlyingThings3D or the fusion of the FlyingThings3D and CityScapes datasets. The ``poly'' learning rate policy is adopted in which the current learning rate equals to $base\_lr\times(1-\frac{iter}{max\_iter})^{power}$. It's demonstrated in \cite{yang_segstereo,zhao2017pyramid} that such learning policy leads to better performance for semantic segmentation and stereo matching tasks. We also adopt SGD for optimization with a minibatch of $16$ stereo pairs. We set the base learning rate to $0.01$, power to $0.9$, momentum to $0.9$ and weight decay to $0.0001$. We run SGD for $200K$ iterations totally in the second stage. For data augmentation, we adopt the random scaling, color shift and contrast adjustment. The random scaling factor is between $0.5$ and $2.0$, the maximum color shift is set to $20$ for each channel and the contrast multiplier is between $0.8$ and $1.2$. We randomly crop $513\times321$ patches for training.

In the third training stage, we pretrain the \emph{EdgeStereo} network using the same training data as the second stage. Except the base learning rate is set to $0.002$, other hyper-parameters in the second stage are kept.

When finetuning on the KITTI datasets, we use the collaboratively pretrained (FlyingThings3D + CityScapes) model from the third training stage. We set the maximum iteration to $50K$ and base learning rate to $0.002$. To prepare for KITTI benchmark submissions, we prolong the second training stage to $500K$ iterations, and adopt the whole training set in the KITTI 2012 or 2015 dataset for finetuning. Since ground-truth disparities are sparse in the KITTI training sets, invalid pixels are neglected in the disparity regression loss.

When finetuning on the Middlebury dataset, we use the Flyingthings3D pretrained model from the third training stage. We collect $35$ image pairs from the Middlebury 2003, 2005, 2006 and 2014 datasets for finetuning, and leave $15$ image pairs from the official Middlebury training set for validation. We set the base learning rate to $0.002$, the batch size to $16$ and the maximum iterations to $10K$. For submission, we use all $50$ image pairs from the training and validation sets and finetune the pretrained model for $10K$ iterations.

The testing is conducted on a single Nvidia GTX 1080Ti GPU. For evaluation on the FlyingThings3D test set, KITTI validation sets and Middlebury dataset, the input sizes are $961\times545$, $1281\times385$ and $897\times601$ respectively. For KITTI submissions, the input size is slightly enlarged to $1313\times393$ for better performance.

For ablation studies, we adopt the disparity estimation sub-network in \emph{EdgeStereo} as the \textbf{baseline} model, without the incorporation of edge cues. The network structure of the baseline model can be easily inferred from Table \ref{t1}.

\subsection{Ablation Studies}

\subsubsection{Ablation Study of Edge Cues}


The first experiment in Table \ref{t2} demonstrates that incorporating edge cues significantly improves the accuracy of disparity estimation. As can be seen, when regularizing the baseline model under the guidance of the edge-aware smoothness loss, the $3$-pixel error on the KITTI 2012 validation set is reduced from $2.844\%$ to $2.555\%$. After embedding the edge features with fine-grained information, the error rate is further reduced to $2.385\%$. 

The second experiment in table \ref{t2} shows that the proposed edge-aware smoothness loss is superior to other smoothness regularization terms. We adopt two sophisticated edge-preserving smoothness terms for comparison: 1) Charbonnier smoothness loss \cite{yang_segstereo} $L_{sm}=\frac{1}{N}\sum_{i,j}[\rho_{s}(d_{i,j}-d_{i+1,j})+\rho_{s}(d_{i,j}-d_{i,j+1})]$, where
$\rho_{s}$ is implemented as the generalized Charbonnier function \cite{barron2017more}; 2) Second-order gradient smoothness loss \cite{zhong2017self} $L_{sm}=\frac{1}{N}\sum_{i,j}|\partial_x^{2}d_{i,j}|e^{-|\partial_{x}^{2}I_{i,j}|}+|\partial_y^{2}d_{i,j}|e^{-|\partial_{y}^{2}I_{i,j}|}$, where $\partial^{2}I$ denotes the second-order image gradient. As can be seen, the edge-aware smoothness loss is more semantic-meaningful and can bring beneficial geometric constraints to disparity estimation, hence achieving the best performance on three datasets compared with other image gradient based regularization terms. In addition, the edge-aware smoothness loss can guide the multi-task learning in \emph{EdgeStereo} and help refine edge predictions, while other smoothness regularization terms can not.

\begin{table}[tb]
\scriptsize
\renewcommand{\captionfont}{\scriptsize}
\centering
\caption{Ablation study of the residual pyramid. EPE on the FlyingThings3D test set is only calculated for pixels whose ground-truth disparity $<192$. }
\label{t3}
\begin{tabularx}{6.7cm}{p{1cm}|p{1.5cm}<{\centering}|X<{\centering}X<{\centering}}
\hlinew{0.75pt}
\multirow{2}{*}{Model} & FlyingThings3D & \multicolumn{2}{c}{KITTI 2012 val} \\
\cline{2-4}
 & EPE & $>3\,px$  & EPE \\
\hline
$no\_RP$ & 0.830 & 2.484 & 0.611 \\
$RP\_8$ & 0.763 & 2.378 & 0.561 \\
$RP\_4$ & \textbf{0.740} & \textbf{2.289} & \textbf{0.542}  \\
$RP\_2$ & 0.751 & 2.385 & 0.555 \\
$cascade$ & 0.772  &  2.407& 0.560 \\
\hlinew{0.75pt}
\end{tabularx}
\end{table}

\begin{table}[tb]
\scriptsize
\renewcommand{\captionfont}{\scriptsize}
\centering
\caption{Ablation study of the edge-aware smoothness loss intensity, which is controlled by $\beta$ in Eq. (\ref{smoo}). Experiments are conducted on the KITTI 2012 validation set.}
\label{t4}
\begin{tabularx}{7.2cm}{p{1.3cm}|X<{\centering}|X<{\centering}|X<{\centering}|X<{\centering}|X<{\centering}}
\hlinew{0.75pt}
$\beta$ & 0 & 1 &2 & 4 &8 \\
\hline
$3$-pixel error & 2.569 & 2.367 & \textbf{2.289} & 2.337 & 2.341 \\
EPE & 0.620 & 0.556 & \textbf{0.542} & 0.546 & 0.552 \\
\hlinew{0.75pt}
\end{tabularx}
\end{table}

\subsubsection{Ablation Study of the Residual Pyramid} To verify the effectiveness of our proposed residual pyramid, we compare three networks containing different residual pyramids ($RP\_2$, $RP\_4$ and $RP\_8$), with an \emph{EdgeStereo} model without the residual pyramid denoted as $no\_RP$, and an \emph{EdgeStereo} model with another cascaded network for refinement denoted as $cascade$. In $no\_RP$, the decoder consists of $3$ deconvolutional blocks and $1$ convolutional layer to produce a full-size disparity map. In $cascade$, similar to iResNet \cite{liang2017learning}, an additional network is cascaded on $no\_RP$ to predict full-size residual signals.

As listed in Table \ref{t3}, three residual pyramids are all helpful for disparity estimation, because of the leveraged knowledge from stereo geometry and geometric constraints. In addition, compared with disparity refinement using another cascaded network \cite{liang2017learning}, the coarse-to-fine residual learning mechanism alleviates the
problem of over-fitting and achieves better performance. The best performing $RP\_4$ yields an error rate of $2.289\%$ on the KITTI 2012 validation set and an EPE of $0.740$ on the FlyingThings3D test set, which is adopted in the final \emph{EdgeStereo} model.

\subsubsection{Ablation Study of the Edge-aware Smoothness Loss Intensity}

In Eq. (\ref{smoo}), the hyper-parameter $\beta$ controls the intensity of the edge-aware smoothness loss. As shown in Table \ref{t4}, when $\beta$ is zero, the edge-aware smoothness loss is degraded to a simple $L1$ regularization term, which will over-smooth disparity maps; when $\beta$ is large, the smoothness regularization term is sensitive to noises in edge estimates, hence harming the performance of disparity estimation. The best setting ($\beta=2$) yields an error rate of $2.289\%$ on the KITTI 2012 validation set, which is kept in the following experiments.

\subsection{Effectiveness of Multi-task Learning}

\begin{table}[tb]
\scriptsize
\renewcommand{\captionfont}{\scriptsize}
\centering
\caption{Quantitative demonstrations of better edge predictions after multi-task learning. Experiments are conducted on the BSDS500 training, val and test sets.}
\label{t5}
\begin{tabularx}{8.8cm}{p{3.7cm}|X<{\centering}X<{\centering}|X<{\centering}X<{\centering}|X<{\centering}X<{\centering}}
\hlinew{0.75pt}
\multirow{2}{*}{Model} & \multicolumn{2}{c|}{training} & \multicolumn{2}{c|}{val} & \multicolumn{2}{c}{test} \\
\cline{2-7}
 & ODS & OIS & ODS & OIS & ODS & OIS \\
\hline
Multicue-pretrained edge branch & 0.328 & 0.368 & 0.332 & 0.373 & 0.340 & 0.373 \\
Edge branch after multi-task learning & \textbf{0.455} & \textbf{0.477} & \textbf{0.458} & \textbf{0.476} & \textbf{0.460} & \textbf{0.476} \\
\hlinew{0.75pt}
\end{tabularx}
\end{table}

\begin{figure*}[tb]
\centering
\includegraphics[width=1\textwidth]{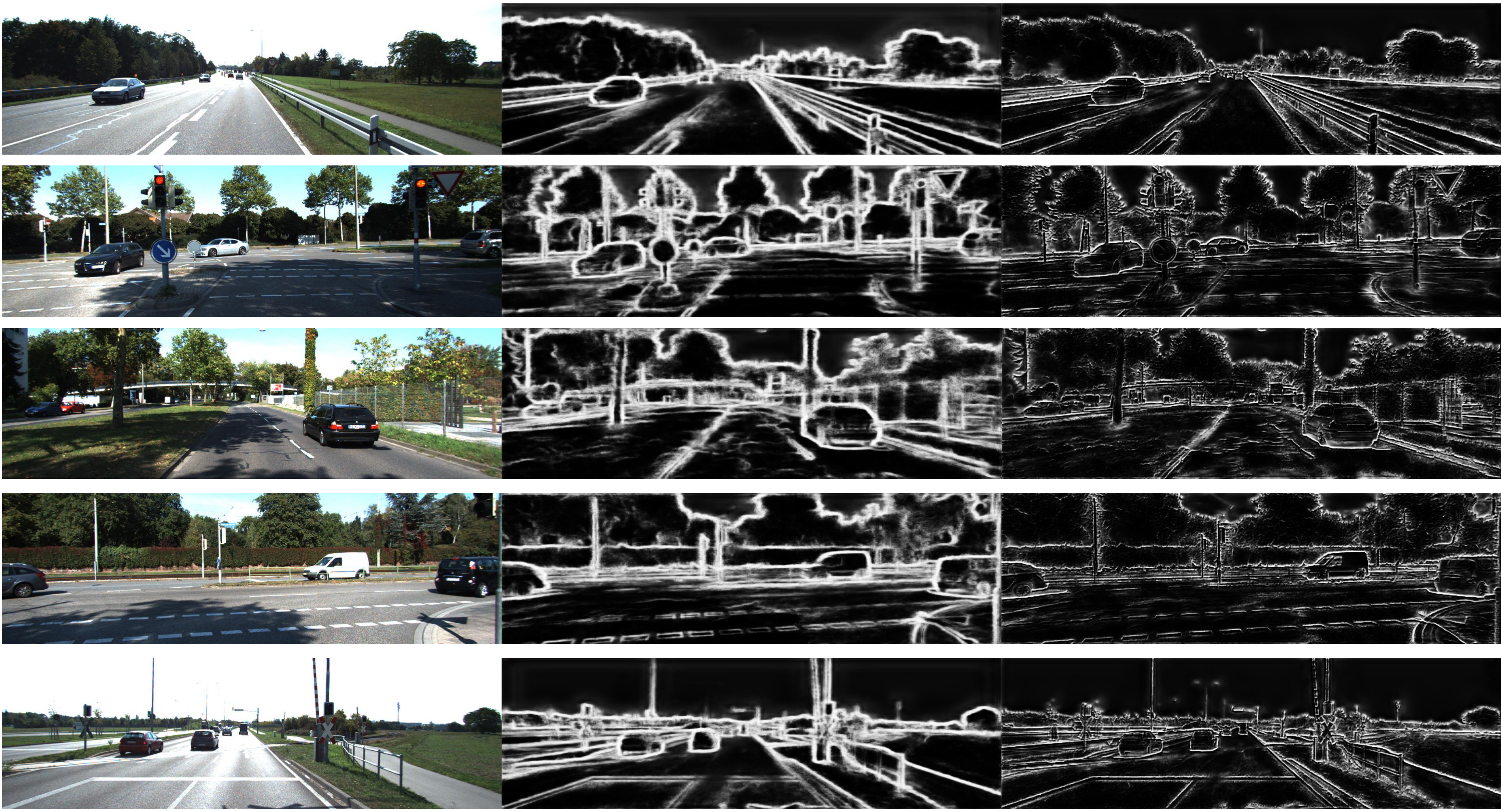}
\caption{Qualitative demonstrations of better edge predictions after multi-task learning. From left: image in the KITTI 2015 training set, edge map from the Multicue-pretrained edge detection sub-network, edge map from \emph{EdgeStereo} after multi-task learning. Non-maximum suppression (NMS) is \textbf{not} adopted to thin detected edges. }
\label{edge}
\end{figure*}

\begin{table}[tb]
\scriptsize
\renewcommand{\captionfont}{\scriptsize}
\centering
\caption{Quantitative demonstrations of better disparity estimates near boundaries after incorporating edge cues. Experiments are conducted on the KITTI validation sets.}
\label{t6}
\begin{tabularx}{8.8cm}{p{3.5cm}|X<{\centering}X<{\centering}|X<{\centering}X<{\centering}}
\hlinew{0.75pt}
\multirow{2}{*}{Model} & \multicolumn{2}{c|}{KITTI 2012 val} & \multicolumn{2}{c}{KITTI 2015 val}  \\
\cline{2-5}
 & $>3\,px$  & EPE & $>3\,px$  & EPE  \\
\hline
Baseline model (without edge cues) & 3.957 & 0.764 & 4.354 & 0.989  \\
\emph{EdgeStereo} & \textbf{3.401} & \textbf{0.704} & \textbf{3.968} & \textbf{0.946} \\
\hlinew{0.75pt}
\end{tabularx}
\end{table}

\begin{table*}[tb]
\scriptsize
\renewcommand{\captionfont}{\scriptsize}
\centering
\caption{Comparison with other stereo matching methods on the FlyingThings3D test set.}
\label{t7}
\begin{tabularx}{1\textwidth}{p{0.9cm}|p{0.65cm}<{\centering}|p{0.75cm}<{\centering}|p{1.4cm}<{\centering}|p{1.05cm}<{\centering}|p{0.75cm}<{\centering}|p{1.2cm}<{\centering}|X<{\centering}|X<{\centering}|X<{\centering}|p{0.9cm}<{\centering}|X<{\centering}}
\hlinew{0.75pt}
\multicolumn{2}{c|}{Metric} & SGM\cite{hirschmuller2005accurate} & MC-CNN\cite{zbontar2016stereo} &  DispNet\cite{mayer2016large} & CRL\cite{pang2017cascade} & GC-Net\cite{kendall2017end} & PSMNet\cite{chang2018pyramid} & iResNet\cite{liang2017learning} & SegStereo\cite{yang_segstereo} & Baseline & \emph{EdgeStereo} \\
\hline
\multirow{2}{*}{Protocol 1} & $>3\,px$ & 12.54 & 13.70  & 9.67 & 6.37 & 8.13 & 4.69 & 4.58 & 4.61 & 4.72 & \textbf{4.30} \\
& EPE & 4.50 & 3.79  & 1.84 & 1.33 & 2.83 & 1.65 & \textbf{1.05} & 1.36 & 1.32 & 1.27 \\
\hline
\multirow{2}{*}{Protocol 2} & $>3\,px$ & \verb|-| & \verb|-|  & 9.31 & 5.97 & 7.73 & 4.14 & 4.40 & 4.27 & 4.40 &\textbf{3.96} \\
& EPE & \verb|-| & \verb|-|  & 1.70 &1.21 & 2.15 & 0.98 & 0.95 & 0.91 & 0.82 & \textbf{0.74}  \\
\hlinew{0.75pt}
\end{tabularx}
\end{table*}

\begin{table*}[tb]
\scriptsize
\renewcommand{\captionfont}{\scriptsize}
\centering
\caption{Comparison with other stereo matching methods in  \textbf{``Reflective Regions''} on the KITTI stereo 2012 benchmark (March 5, 2019).}
\label{t8}
\begin{tabularx}{1\textwidth}{p{0.7cm}<{\centering}|p{0.75cm}<{\centering}|p{1.4cm}<{\centering}|X<{\centering}|p{1.2cm}<{\centering}|X<{\centering}|X<{\centering}|X<{\centering}|X<{\centering}|p{1.2cm}<{\centering}|p{0.75cm}<{\centering}|p{1.0cm}<{\centering}}
\hlinew{0.75pt}
Metric & SGM\cite{hirschmuller2005accurate}& MC-CNN\cite{zbontar2016stereo} & DispNet\cite{mayer2016large} & GC-Net\cite{kendall2017end} & Displets\cite{guney2015displets} & PSMNet\cite{chang2018pyramid} & iResNet\cite{liang2017learning} & PDSNet\cite{tulyakov2018practical} & SegStereo\cite{yang_segstereo}  & Baseline & \emph{EdgeStereo} \\
\hline
$>3\,px$ & 27.39 & 17.09 & 16.04 & 10.80 & 8.40 & 8.36 & 7.40 & 6.50 & 6.35 & 8.53 & \textbf{5.84}  \\
EPE & 5.1 px & 3.2 px & 2.1 px & 1.8 px & 1.9 px & 1.4 px & 1.2 px & 1.4 px & 1.1 px & 1.5 px & \textbf{1.0 px} \\

\hlinew{0.75pt}
\end{tabularx}
\end{table*}

\subsubsection{Stereo Matching Helps Edge Detection} \emph{EdgeStereo\_RP\_4} is adopted for verification. We first train the edge branch on Multicue, then fix it and train the disparity branch on FlyingThings3D, finally train two branches simultaneously on FlyingThings3D. We compare the Multicue-pretrained edge detection sub-network with the edge branch in \emph{EdgeStereo} after multi-task learning.

Firstly we give quantitative demonstrations on the BSDS500 edge detection dataset. As shown in Table \ref{t5}, ODS F-measures and OIS F-measures on the BSDS500 training, validation and test sets are all improved after multi-task learning, even though the BSDS500 dataset is not used for pretraining and the FlyingThings3D training set does not contain ground-truth edge annotations during training.

Next we give qualitative demonstrations on the KITTI stereo 2015 training set without edge annotations. As shown in Fig. \ref{edge}, after multi-task learning, edge predictions are significantly refined and details are highlighted in the produced edge maps, even though the model is not trained on KITTI 2015. Hence the mutual exploitation of stereo and edge information under the guidance of the edge-aware smoothness loss is beneficial for edge detection task, proving the effectiveness of the multi-task learning in our unified model.

\subsubsection{Edge Detection Helps Stereo Matching}
Several qualitative and quantitative demonstrations are already provided in Fig. \ref{intro} and Table \ref{t1}. We further prove that disparity estimates near boundaries are more accurate after incorporating edge cues. Considering edge annotations are not provided in the KITTI stereo datasets, we treat the edge predictions from the FlyingThings3D-pretrained \emph{EdgeStereo} model as boundaries for evaluation. As shown in Table \ref{t6}, after incorporating edge cues, $3$-pixel errors near boundaries are reduced by $14.1\%$ and $8.7\%$ on the KITTI 2012 and 2015 validation sets respectively, compared with the baseline model.

\begin{table*}[tb]
\scriptsize
\renewcommand{\captionfont}{\scriptsize}
\centering
\caption{Comparison with other stereo matching methods on the KITTI stereo 2012 benchmark (March 1, 2019).}
\label{t9}
\begin{tabularx}{13.5cm}{p{2.6cm}|X<{\centering}X<{\centering}|X<{\centering}X<{\centering}|X<{\centering}X<{\centering}|X<{\centering}X<{\centering}|X<{\centering}}
\hlinew{0.75pt}
& \multicolumn{2}{c|}{$>\,2$px} & \multicolumn{2}{c|}{$>\,3$px} & \multicolumn{2}{c|}{$>\,4$px} & \multicolumn{2}{c|}{$>\,5$px} & EPE\\
& Noc & All & \textbf{\emph{Noc}} & All & Noc & All  & Noc & All & Noc \\
\hline
PSMNet \cite{chang2018pyramid} & 2.44 & 3.01 & 1.49 & 1.89  & 1.12 & 1.42  & 0.90 & 1.15 & 0.5 px \\
SegStereo\cite{yang_segstereo} & 2.66 & 3.19 & 1.68 & 2.03 & 1.25 & 1.52 & 1.00 & 1.21 & 0.5 px \\
iResNet \cite{liang2017learning} & 2.69 & 3.34 & 1.71 & 2.16  & 1.30 & 1.63  & 1.06 & 1.32 & 0.5 px \\
GC-Net \cite{kendall2017end} & 2.71 & 3.46 & 1.77 & 2.30 & 1.36 & 1.77  & 1.12 & 1.46 & 0.6 px \\
PDSNet \cite{tulyakov2018practical} & 3.82 & 4.64 & 1.92 & 2.53 & 1.38 & 1.85 & 1.12 & 1.51 & 0.9 px \\
L-ResMatch \cite{shaked2016improved} & 3.64 & 5.06 & 2.27 & 3.40 & 1.76 & 2.67 & 1.50 & 2.26 & 0.7 px \\
SGM-Net \cite{seki2017sgm} & 3.60 & 5.15 & 2.29 & 3.50  & 1.83 & 2.80  & 1.60 & 2.36 & 0.7 px\\
Displets \cite{guney2015displets} & 3.90 & 4.92 & 2.37 & 3.09  & 1.97 & 2.52 & 1.72 & 2.17 & 0.7 px\\
MC-CNN \cite{zbontar2016stereo}& 3.90 & 5.45 & 2.43 & 3.63  & 1.90 & 2.85  & 1.64 & 2.39 & 0.7 px\\
DispNet \cite{mayer2016large} & 7.38 & 8.11 & 4.11 & 4.65 & 2.77 & 3.20 & 2.05 & 2.39 & 0.9 px\\
SGM \cite{hirschmuller2005accurate} & 8.66 & 10.16 & 5.76 & 7.00 & 4.38 & 5.41 & 3.56 & 4.41 & 1.2 px\\
\hline
Baseline & 2.91 & 3.55 & 1.81 & 2.27 & 1.29 & 1.61 & 1.07 & 1.25 & 0.5 px \\
\emph{EdgeStereo} & \textbf{2.32} & \textbf{2.88}  & \textbf{1.46} & \textbf{1.83}  & \textbf{1.07} & \textbf{1.34}  & \textbf{0.83} & \textbf{1.04} & \textbf{0.4 px}\\
\hlinew{0.75pt}
\end{tabularx}
\end{table*}

\begin{table*}[tb]
\scriptsize
\renewcommand{\captionfont}{\scriptsize}
\centering
\caption{Comparison with other stereo matching methods on the KITTI stereo 2015 benchmark (March 1, 2019).}
\label{t10}
\begin{tabularx}{13.5cm}{p{2.6cm}|X<{\centering}X<{\centering}X<{\centering}|X<{\centering}X<{\centering}X<{\centering}|X<{\centering}}
\hlinew{0.75pt}
& \multicolumn{3}{c|}{All Pixels} & \multicolumn{3}{c|}{Non-Occluded Pixels} & Runtime \\
& D1-bg & D1-fg & \textbf{\emph{D1-all}} & D1-bg & D1-fg & D1-all & (s) \\
\hline
SegStereo\cite{yang_segstereo} & 1.88 & 4.07 & 2.25 & 1.76 & 3.70 & 2.08 & 0.6 \\
PSMNet \cite{chang2018pyramid} & 1.86 & 4.62 & 2.32 & 1.71 & 4.31 & 2.14 & 0.41\\
iResNet \cite{liang2017learning} & 2.25 & 3.40 & 2.44 & 2.07 & \textbf{2.76} & 2.19 & 0.12\\
PDSNet \cite{tulyakov2018practical} & 2.29 & 4.05 & 2.58 & 2.09 & 3.68 & 2.36 & 0.5\\
CRL \cite{pang2017cascade} & 2.48 & 3.59 & 2.67 & 2.32 & 3.12 & 2.45 & 0.47\\
GC-Net \cite{kendall2017end} & 2.21 & 6.16 & 2.87 & 2.02 & 5.58 & 2.61 & 0.9\\
LRCR \cite{jie2018left} & 2.55 & 5.42 & 3.03 & 2.23 & 4.19 & 2.55 & 49.2\\
DRR \cite{gidaris2016detect} & 2.58 & 6.04 & 3.16 & 2.34 & 4.87 & 2.76 & 0.4 \\
L-ResMatch \cite{shaked2016improved} & 2.72 & 6.95 & 3.42 & 2.35 & 5.74 & 2.91 & 48\\
Displets \cite{guney2015displets} & 3.00 & 5.56 & 3.43 & 2.73 & 4.95 & 3.09 & 265\\
SGM-Net \cite{seki2017sgm} & 2.66 & 8.64 & 3.66 & 2.23 & 7.44 & 3.09 & 67\\
MC-CNN \cite{zbontar2016stereo} & 2.89 & 8.88 & 3.88 & 2.48 & 7.64 & 3.33 & 67\\
DispNet \cite{mayer2016large} & 4.32 & 4.41 & 4.34 & 4.11 & 3.72 & 4.05 & \textbf{0.06}\\
SGM \cite{hirschmuller2005accurate} & 5.06 & 13.00 & 6.38 & 4.43 & 11.68 & 5.62 & 0.11 \\
\hline
Baseline & 2.11 & 3.99 & 2.41 & 1.94 & 3.35 & 2.17 & 0.26 \\
\emph{EdgeStereo}  & \textbf{1.84} & \textbf{3.30} & \textbf{2.08} & \textbf{1.69} & 2.94 & \textbf{1.89} &  0.32\\
\hlinew{0.75pt}
\end{tabularx}
\end{table*}

\begin{table*}[htb]
\scriptsize
\renewcommand{\captionfont}{\scriptsize}
\centering
\caption{Comparison with other stereo matching methods on the Middlebury benchmark (July 29, 2019). }
\label{tmid}
\begin{tabularx}{12.7cm}{p{2cm}|p{2cm}<{\centering}|X<{\centering}|X<{\centering}|p{2cm}<{\centering}|p{1.65cm}<{\centering}}
\hlinew{0.75pt}
  & MC-CNN-fst\cite{zbontar2016stereo} (Half) & SGM\cite{hirschmuller2005accurate} (Half) & iResNet\cite{liang2017learning} (Half) & PSMNet\_ROB\cite{chang2018pyramid} (Quarter) & EdgeStereo (Full) \\
\hline
$2$-pixel error & \textbf{9.47\%} & 18.4\% & 22.9\% & 42.1\% & 18.7\% \\
\hlinew{0.75pt}
\end{tabularx}
\end{table*}

\subsection{Comparison with Other Stereo Matching Methods}

\emph{EdgeStereo\_RP\_4} is adopted as the final \emph{EdgeStereo} model, with three outputs for training. Through experiments, the loss weights $\lambda_{r}^{2},\lambda_{r}^{1},\lambda_{r}^{0}$ for the disparity regression loss are set to $0.6,0.8,1.0$; the loss weights $\lambda_{sm}^{2},\lambda_{sm}^{1},\lambda_{sm}^{0}$ for the edge-aware smoothness loss are set to $0.06,0.08,0.1$.

\subsubsection{FlyingThings3D Results} We first compare \emph{EdgeStereo} with several non-end-to-end stereo matching methods, including SGM \cite{hirschmuller2005accurate}, MC-CNN \cite{zbontar2016stereo} and DRR \cite{gidaris2016detect}. Next we compare \emph{EdgeStereo} with the state-of-the-art end-to-end stereo matching networks, including DispNetC \cite{mayer2016large}, CRL \cite{pang2017cascade}, GC-Net \cite{kendall2017end}, PSMNet \cite{chang2018pyramid}, iResNet \cite{liang2017learning} and SegStereo \cite{yang_segstereo}. For comparison, we use models made available by authors except GC-Net, while we retrain GC-Net following the settings in their paper. As can be seen in Table \ref{t7}, \emph{EdgeStereo} achieves the best performance under two testing protocols.

\begin{figure*}[tb]
\centering
\includegraphics[width=1\textwidth]{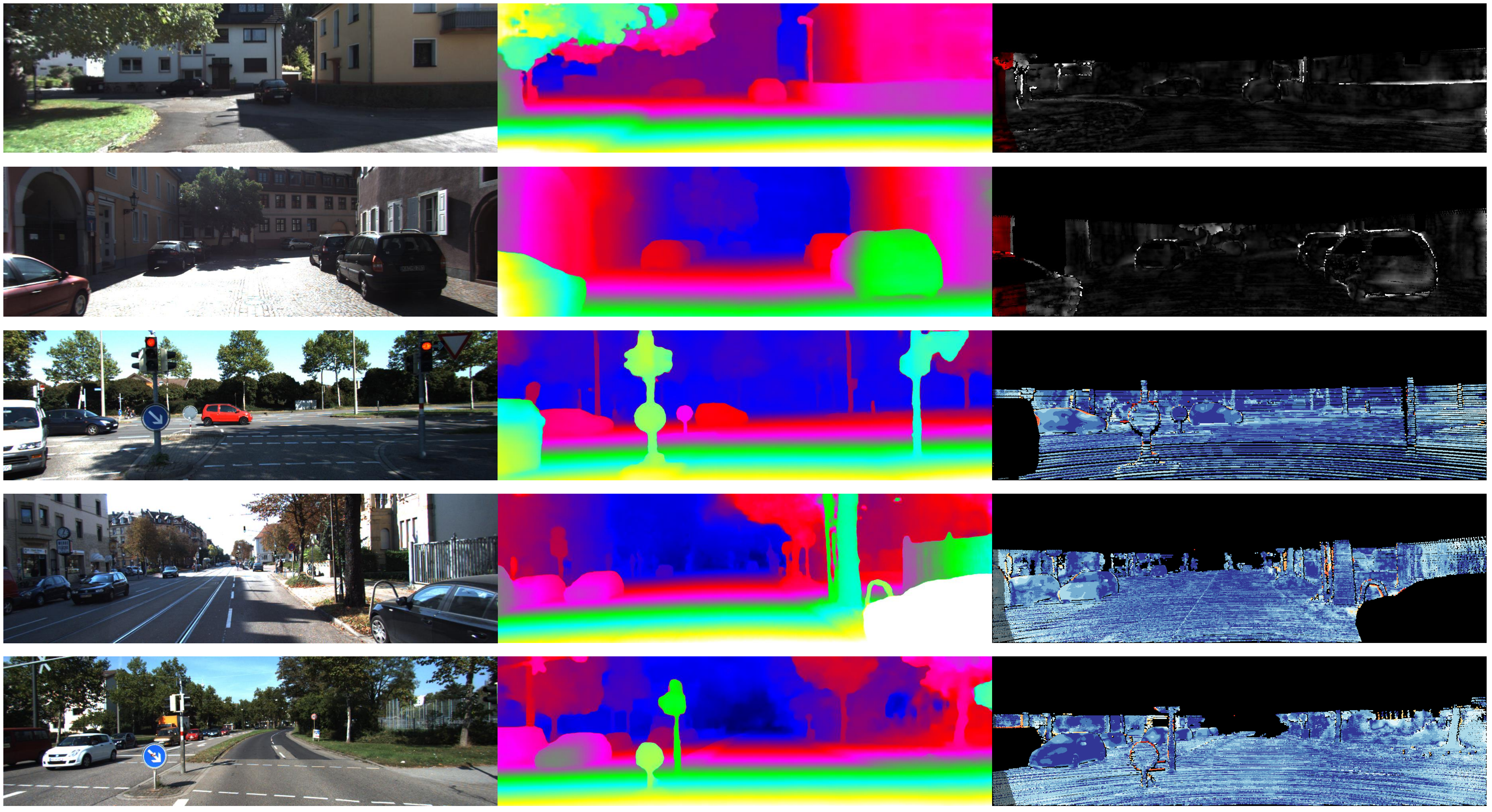}
\caption{Qualitative results on the KITTI 2012 and 2015 test sets. From left: left stereo image, disparity estimates, error map.}
\label{kitti}
\end{figure*}

\begin{table*}[tb]
\scriptsize
\renewcommand{\captionfont}{\scriptsize}
\centering
\caption{The generalization performance of state-of-the-art end-to-end stereo matching methods. To compare generalization capability, FlyingThings3D-pretrained models are evaluated on the KITTI 2012 and 2015 training sets, collaborative-pretrained (FlyingThings3D plus CityScapes) models are evaluated on the KITTI 2012 and Middlebury 2014 training sets.}
\label{t11}
\begin{tabularx}{1\textwidth}{p{2.5cm}|p{0.9cm}<{\centering}|X<{\centering}|X<{\centering}|X<{\centering}|X<{\centering}|X<{\centering}|X<{\centering}|X<{\centering}}
\hlinew{0.75pt}
\multicolumn{2}{c|}{Dataset and Metric} & DispNet & CRL & iResNet & PSMNet & SegStereo & Baseline & \emph{EdgeStereo} \\
\hline
\multicolumn{9}{c}{\emph{1. FlyingThings3D Pretrain}} \\
\hline
\multirow{2}{*}{KITTI 2012 training} & $>3\,px$ & 12.542 & 9.073 & \textbf{7.895} & 27.333 & 12.808 & 16.783 & 12.274 \\
& EPE & 1.753 & 1.387 & \textbf{1.278} & 5.549  & 2.052 & 3.333 & 1.963 \\
\hline
\multirow{2}{*}{KITTI 2015 training} & $>3\,px$ & 12.881 & 8.885 & \textbf{7.424} & 29.868  & 11.234 & 16.179 & 12.467\\
& EPE & 1.596 & 1.357 & \textbf{1.213} &6.445  & 2.187 & 3.653 & 2.068 \\
\hline
\multicolumn{9}{c}{\emph{2. FlyingThings3D + CityScapes Pretrain}} \\
\hline
\multirow{2}{*}{KITTI 2012 training} & $>3\,px$ & 6.373  &  5.705 & \textbf{4.867} & 5.716  & 5.476 & 5.677 & 5.239 \\
& EPE & 1.194  & 1.096 & 1.001 & 1.137  & 1.113 & 1.010 & \textbf{0.999} \\
\hline
\multirow{2}{*}{Middlebury (Half)} & $>3\,px$ & 15.684 & 10.987 & \textbf{10.475} & 13.859 & 14.477  & 13.120 & 11.136 \\
& EPE & 2.217  & 1.676 & 1.583 & 1.987  & 2.173 & 2.109 & \textbf{1.552} \\
\hlinew{0.75pt}
\end{tabularx}
\end{table*}

\subsubsection{KITTI 2012 Results}

\emph{EdgeStereo} is finetuned using all $194$ training pairs, then the testing results are submitted to the KITTI 2012 online leaderboard. For evaluation, we use the percentage of erroneous pixels ($>2px$, $>3px$, $>4px$, $>5px$) and EPE in non-occluded (Noc) and all (All) regions. The results are shown in Table \ref{t9}. As can be seen, by the time of the paper submission, \emph{EdgeStereo} outperforms all published stereo matching methods in all evaluation metrics. We also finetune the baseline model on the KITTI 2012 training set, then submit corresponding results to the benchmark. By leveraging edge cues, \emph{EdgeStereo} is obviously superior to the baseline model, producing more reliable disparity estimates in detailed structures, large occlusions and near boundaries.

We also compare \emph{EdgeStereo} with state-of-the-art stereo methods in \textbf{``Reflective Regions''} on the KITTI stereo 2012 benchmark. As shown in Table \ref{t8}, \emph{EdgeStereo} surpasses the baseline model and other methods by a noteworthy margin, especially the SegStereo \cite{yang_segstereo} which utilizes foreground semantic information, and the Displets \cite{guney2015displets} which resolves stereo ambiguities using object knowledge. Hence incorporating semantic-meaningful edge information can provide beneficial geometric knowledge for finding correspondences in texture-less regions.

\subsubsection{KITTI 2015 Results}

We also submit the testing results to the KITTI 2015 online leaderboard. For evaluation, we use the $3$-pixel error of background (D1-bg), foreground (D1-fg) and all pixels (D1-all) in non-occluded and all regions. The results are shown in Table \ref{t10}. As can be seen, \emph{EdgeStereo} achieves the best performance compared with the baseline model and all published stereo matching methods, meanwhile it is more efficient than the 3-D convolutional neural networks and cascaded structures. Fig. \ref{kitti} gives qualitative results on the KITTI test sets. 

\subsubsection{Middlebury 2014 Results}

On the validation set (half size), the baseline model without edge cues achieves a a $2$-pixel error of $12.473\%$ and an EPE of $1.225$, while \emph{EdgeStereo} achieves a $2$-pixel error of $11.540\%$ and an EPE of $1.139$, demonstrating the effectiveness of incorporating edge cues for disparity learning.

Next we compare \emph{EdgeStereo} with other methods on the benchmark. Since the Middlebury dataset is too small, among published end-to-end stereo networks, only iResNet and PSMNet (ROB) report their results on this tiny dataset. As can be seen from Table \ref{tmid}, \emph{EdgeStereo} outperforms iResNet and PSMNet by a noteworthy margin. However it performs worse than the non-end-to-end method MC-CNN. On the Middlebury benchmark, the top-performing methods are non-end-to-end networks and some sophisticated hand engineered methods rather than end-to-end stereo networks. For non-end-to-end networks, training is conducted based on pairs of image patches and the Middlebury training set can provide sufficient training samples, hence their training processes are much more complete than end-to-end networks. The proposed \emph{EdgeStereo} is an end-to-end architecture with a large capacity of feature representation and context characterization. Even if powerful data augmentations are used during training, the Middlebury training set is too small to fit the capacity of \emph{EdgeStereo}. Hence we argue that the powerfulness of \emph{EdgeStereo} should be better evaluated on datasets with larger scales.

\subsection{Generalization Performance}

\subsubsection{Cross-domain Experiments}

The generalization capability of an end-to-end disparity estimation network is important, since obtaining ground-truth disparities using LiDAR is costly, and most real-world stereo datasets are not large enough to train a model without over-fitting. FlyingThings3D pretraining and collaborative pretraining \cite{yang2018srcdisp} (FlyingThings3D plus CityScapes) are the most effective pretraining methods for end-to-end stereo networks, which are adopted to compare the generalization performance of state-of-the-art stereo matching methods. Firstly, we compare FlyingThings3D pretrained models on the KITTI 2012 and 2015 training sets. We use publicly available models (provided by authors) of DispNet, CRL, SegStereo and PSMNet for comparison and train iResNet on Flyingthings3D. Secondly, we compare collaboratively pretrained (FlyingThings3D plus CityScapes) models on the KITTI 2012 and Middlebury 2014 training sets. Since the collaborative pretraining scheme is not used in the previously published methods, we pretrain all competitors (DispNet, CRL, SegStereo, iResNet and PSMNet) following the same scheduling reported in the original papers.

The results are shown in Table \ref{t11}. As can be seen, Flyingthings3D pretrained \emph{EdgeStereo} model achieves comparable generalization performance with DispNet and SegStereo, meanwhile being significantly better than 3-D convolutional networks (PSMNet). However, Flyingthings3D pretrained \emph{EdgeStereo} preforms slightly worse than CRL and iResNet on KITTI training sets, since the capacity of our \emph{EdgeStereo} is much larger than CRL and iResNet, while the synthetic Flyingthings3D dataset has a quite different domain with real-world KITTI datasets (\emph{e.g.} unreasonably large disparities) that \emph{EdgeStereo} has learnt to adapt. When comparing collaboratively pretrained models, where the real-world Cityscapes dataset is used for pretraining, \emph{EdgeStereo} achieves comparable generalization performance with iResNet. In addition, \emph{EdgeStereo} outperforms all other competitors on the Middlebury benchmark, KITTI benchmarks and Flyingthings3D test set, demonstrating the capability of our \emph{EdgeStereo} on different domains. Collaboratively pretrained \emph{EdgeStereo} achieves a $3$-pixel error of $5.239\%$ on the KITTI 2012 training set, which is a promising result for practical use when obtaining dense ground-truth disparities is costly. In addition, when pretrained and evaluated on datasets with different domains, \emph{EdgeStereo} outperforms the baseline model obviously after incorporating edge cues for multi-task learning.

\begin{figure*}[tb]
\centering
\caption{Disparity estimations of the collaboratively pretrained \emph{EdgeStereo} on the self-built Middlebury validation set (indoor edgeless scenarios).}
\vspace{-3mm}
\includegraphics[width=1\textwidth]{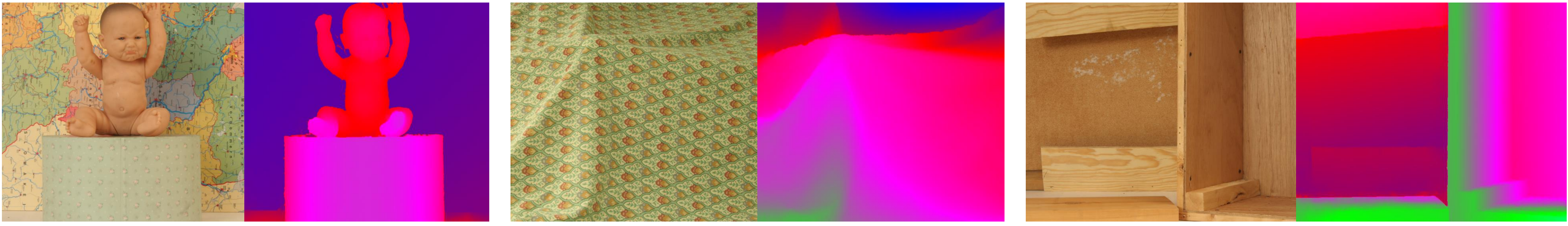}
\label{fmid}
\vspace{-2mm}
\end{figure*}

\begin{table*}[tb]
\scriptsize
\renewcommand{\captionfont}{\scriptsize}
\centering
\vspace{-2mm}
\caption{Comparison between the collaboratively pretrained \emph{EdgeStereo} and baseline on the self-built Middlebury validation set (indoor edgeless scenarios). }
\vspace{-3mm}
\label{t12}
\begin{tabularx}{8.7cm}{p{2cm}|X<{\centering}|X<{\centering}}
\hlinew{0.75pt}
  & $3$-pixel error & EPE  \\
\hline
Baseline  & 2.742 & 0.622 \\
EdgeStereo & \textbf{2.252} & \textbf{0.570} \\
\hlinew{0.75pt}
\end{tabularx}
\vspace{-2mm}
\end{table*}

\begin{table*}[tb]
\scriptsize
\renewcommand{\captionfont}{\scriptsize}
\centering
\caption{Comparison between our new work and the previous version \cite{song2018stereo}.}
\vspace{-3mm}
\label{tcom}
\begin{tabularx}{15.2cm}{p{1.5cm}|X<{\centering}|X<{\centering}|X<{\centering}|X<{\centering}}
\hlinew{0.75pt}
  & KITTI2012 test & KITTI2012 reflective regions & KITTI2015 test & Flyingthings3D  \\
\hline
Previous \cite{song2018stereo} & 1.73\% & 7.01\% & 2.59\% & 4.70\%/0.99 \\
New  & \textbf{1.46\%} & \textbf{5.84\%} & \textbf{2.08\%} & \textbf{3.96\%/0.74}  \\
\hlinew{0.75pt}
\end{tabularx}
\vspace{-2mm}
\end{table*}

\subsubsection{Performance in Indoor Edgeless Scenarios}

To further demonstrate that our proposed method also works well in edgeless indoor scenarios, firstly we choose $10$ edgeless stereo pairs from the Middlebury training set \footnote{Baby1\_06, Baby2\_06, Barn1\_01, Barn2\_01, Bull\_01, Cloth1\_06, Poster\_01, Sawtooth\_01, Venus\_01, Wood1\_06} as a validation set, next we compare the collaboratively pretrained (FlyingThings3D plus CityScapes) \emph{EdgeStereo} model and baseline model (without the edge detection branch) on this validation set.

As can be seen from Table \ref{t12}, the \emph{EdgeStereo} model achieves a satisfactory $3$-pixel error of $2.252\%$ on this challenging full-resolution Middlebury validation set, even if the model is pretrained using unrealistic data (Flyingthings3D) and outdoor data (Cityscapes). In addition, \emph{EdgeStereo} outperforms the baseline model on these edgeless scenarios after multi-task learning, demonstrating that the proposed model would not suffer from the edge branch in these scenarios. The qualitative demonstrations are given in Fig. \ref{fmid}. As can be seen, the collaboratively pretrained \emph{EdgeStereo} produces consistent disparity estimations in edgeless scenarios.

\subsection{Comparison with the Previous Version}

Compared with our ACCV version \cite{song2018stereo}: (1) We change the backbone from the ImageNet pretrained VGG16 to a more powerful feature extractor ResNet50, and we also change the encoder structure to a ResNet-like structure for better intermediate representations; (2) We re-design the structure of the residual pyramid and make it more effective, where multi-scale features are used for binocular warping and cost volume generation; (3) We refine multi-task learning mechanism, where the edge-aware smoothness loss is enabled to be propagated back to the disparity branch and edge branch simultaneously, then both the disparity map and edge map can be optimized under the guidance of the edge-aware smoothness loss. Consequently we obtain the most important conclusion in this paper: edge detection and stereo matching can help each other based on our unified model.

In conclusion, we re-design the overall architecture where all modules are proved to be more effective. Finally, we will give some quantitative comparisons in Table \ref{tcom}. As can be seen, the new architecture outperforms the original one on all test sets and the challenging reflective regions.

\section{Conclusion and Future Work}
\label{s5}
In this paper, we propose an effective multi-task learning network \emph{EdgeStereo} for stereo matching and edge detection. To effectively incorporate edge cues into the disparity estimation pipeline for multi-task learning, we design the edge feature embedding and propose the edge-aware smoothness loss. The experimental results show that the disparity estimates in texture-less regions, large occlusions, detailed structures and near boundaries are significantly improved after incorporating edge cues. Correspondingly, our \emph{EdgeStereo} achieves the best performance on the FlyingThings3D dataset, meanwhile outperforming other published stereo matching methods on the KITTI stereo 2012 and 2015 benchmarks. \emph{EdgeStereo} also ranks the first on the online leaderboard of the KITTI 2012 ``Reflective Regions'' evaluation. In addition, we provide both qualitative and quantitative demonstrations that edge predictions are improved after multi-task learning, even if ground-truth edge annotations are not provided for training. Finally we prove that \emph{EdgeStereo} has a comparable generalization capability for disparity estimation. In conclusion, stereo matching task and edge detection task can promote each other through the geometric knowledge learned from the multi-task interactions in \emph{EdgeStereo}.

In future work, we intend to apply the multi-task learning mechanism of \emph{EdgeStereo} to other dense matching applications, such as optical flow and multi-view reconstruction \emph{etc.} In addition, we consider to explicitly model geometric constraints from stereo matching for edge detection to further improve the quality of edge predictions. Finally, we consider to incorporate more tasks, such as semantic segmentation and instance segmentation \emph{etc}, into our unified model. It will be a future direction to achieve an integrated vision system with a clear and interpretable architecture.

\begin{acknowledgements}
We would like to thank Guorun Yang for helpful discussions and suggestions. This research is supported by the funding from NSFC programs (61673269, 61273285, U1764264).
\end{acknowledgements}

\bibliographystyle{spmpsci}      
\bibliography{ref}

\end{document}